\pdfoutput=1
\documentclass[11pt]{article}

\usepackage[final]{acl}

\usepackage{times}
\usepackage{soul}
\usepackage{xcolor}
\definecolor{darkgreen}{RGB}{0,100,0}
\usepackage{latexsym}
\usepackage{tabularx}
\usepackage{booktabs}
\usepackage{amsmath}
\usepackage{enumitem}
\usepackage{makecell}
\usepackage{adjustbox}
\usepackage{multirow,makecell}

\usepackage[table]{xcolor}

\usepackage[T1]{fontenc}
\usepackage[utf8]{inputenc}

\usepackage{microtype}
\usepackage{multirow}
\usepackage[many]{tcolorbox}
\usepackage[ruled,linesnumbered]{algorithm2e}

\usepackage{algpseudocode}
\usepackage{hyperref}
\usepackage{cleveref}
\usepackage{inconsolata}

\usepackage{graphicx}
\usepackage{amsmath,amsfonts,bm}

\def\eqref#1{equation~\ref{#1}}
\def\1{\bm{1}}

\def\vtheta{{\bm{\theta}}}
\def\va{{\bm{a}}}

\def\vx{{\bm{x}}}
\def\vy{{\bm{y}}}

\DeclareMathAlphabet{\mathsfit}{\encodingdefault}{\sfdefault}{m}{sl}
\SetMathAlphabet{\mathsfit}{bold}{\encodingdefault}{\sfdefault}{bx}{n}

\newcommand{\system}[1]{#1\xspace}  % Define the system macro with 1 parameter
\newcommand{\DiscoSG}{\system{DiscoSG}}  % Changed command name to avoid numeric confusion
\newcommand{\DiscoSGRefiner}{\system{DiscoSG-Refiner}}
\newcommand{\StanfordParser}{\system{Stanford-Parser}}
\newcommand{\VGTBase}{\system{VG-T5-base}}

\newcommand{\VGTBaseDirect}{\system{VG-T5-base (Direct)}}
\newcommand{\VGTBaseSentMerge}{\system{VG-T5-base (Sent+Merge)}}

\newcommand{\FACTUALTBaseSentMerge}{\system{FACTUAL-T5-base (Sent+Merge)}}

\newcommand{\FACTUALTDirect}{\system{FACTUAL-T5 (Direct)}}
\newcommand{\FACTUALT}{\system{FACTUAL-T5}}
\newcommand{\FACTUALTBase}{\system{FACTUAL-T5-base}}
\newcommand{\TSGBench}{\system{TSGBench}}

\newcommand{\FlanT}{\system{Flan-T5}}
\newcommand{\FlanTBase}{\system{Flan-T5-base}}
\newcommand{\FlanTLarge}{\system{Flan-T5-large}}

\newcommand{\DiscoSGTFive}{\system{DiscoSG-T5}}
\newcommand{\DiscoSGTFiveBase}{\system{DiscoSG-T5-base}}
\newcommand{\DiscoSGTFiveLarge}{\system{DiscoSG-T5-large}}
\newcommand{\DiscoSGTFiveXL}{\system{DiscoSG-T5-xl}}

\newcommand{\Qwen}{\system{Qwen2.5}}
\newcommand{\DiscoSGQwen}{\system{DiscoSG-Qwen2.5}}
\newcommand{\DiscoSGQwenHalfB}{\system{DiscoSG-Qwen2.5-0.5B}}
\newcommand{\DiscoSGQwenOneFiveB}{\system{DiscoSG-Qwen2.5-1.5B}}
\newcommand{\DiscoSGQwenSevenB}{\system{DiscoSG-Qwen2.5-7B}}

\newcommand{\DiscoSGGPT}{\system{DiscoSG-GPT-4o}}

\newcommand{\PiVe}{\system{PiVe}}
\newcommand{\SelfRefine}{\system{Self-Refine}}
\newcommand{\ProgRefine}{\system{Prog-Refine}}

\newcommand{\GraphRAG}{\system{GraphRAG}}

\newcommand{\DiscoSGRefinerBase}{\system{DiscoSG-Refiner-base}}
\newcommand{\DiscoSGRefinerLarge}{\system{DiscoSG-Refiner-large}}
\newcommand{\DiscoSGRefinerXL}{\system{DiscoSG-Refiner-xl}}

\newcommand{\QwenInstruct}{\system{Qwen2.5-72B-Instruct}}
\newcommand{\LlamaInstruct}{\system{Llama-3.3-70B-Instruct}}
\newcommand{\GPTFour}{\system{GPT-4o}}
\newcommand{\GPTFourOne}{\system{GPT-4.1}}
\newcommand{\GPTFourText}{\system{GPT-4o (text-only)}}
\newcommand{\GPTFourMulti}{\system{GPT-4o (multimodal)}}

\newcommand{\dataset}[1]{#1\xspace}
\newcommand{\DiscoSGDS}{\dataset{DiscoSG-DS}}
\newcommand{\DiscoSGED}{\dataset{DiscoSG-ED}}
\newcommand{\VG}{\dataset{VG}}
\newcommand{\FACTUAL}{\dataset{FACTUAL}}
\newcommand{\DetailCaps}{\dataset{DetailCaps}}
\newcommand{\CapArena}{\dataset{CapArena}}

\newcommand{\SharedGPTV}{\dataset{SharedGPT4V}}
\newcommand{\FOIL}{\dataset{FOIL}}
\newcommand{\DFOIL}{\dataset{D-FOIL}}

\newcommand{\vyO}{\vy^0} % Initial graph
\newcommand{\vyt}{\vy^t} % Graph at step t
\newcommand{\vytnext}{\vy^{t+1}} % Graph at step t+1
\newcommand{\vyT}{\vy^T} % Final graph
\usepackage{arydshln}   % for dashed / dotted rules

\title{DiscoSG: Towards Discourse-Level Text Scene Graph Parsing through Iterative Graph Refinement}

\author{
  Shaoqing Lin$^1$,
  Chong Teng$^{1}$\footnotemark[2],
  Fei Li$^{1}$,
  Donghong Ji$^1$, \\
  \textbf{Lizhen Qu$^2$, Zhuang Li$^3$\footnotemark[3]} \\
  $^1$Key Laboratory of Aerospace Information Security and Trusted Computing, \\
  Ministry of Education, School of Cyber Science and Engineering, Wuhan University \\
  $^2$Faculty of Information Technology, Monash University \\
  $^3$School of Computing Technologies, RMIT University \\
  \texttt{\{sqlinn,tengchong,lifei\_csnlp,dhji\}@whu.edu.cn} \\
  \texttt{lizhen.qu@monash.edu} \quad \texttt{zhuang.li@rmit.edu.au}
}

\begin{document}
%  \abovedisplayskip=0.01pt
% \abovedisplayshortskip=0.01pt
% \belowdisplayskip=0.01pt
% \belowdisplayshortskip=0.01pt
\maketitle
\begingroup
\renewcommand{\thefootnote}{\fnsymbol{footnote}}
\footnotetext[2]{Corresponding author.}  % † (for Chong Teng)
\footnotetext[3]{Senior author.}         % ‡ (for Zhuang Li)
\endgroup
\begin{abstract}
Vision-Language Models (VLMs) generate discourse-level, multi-sentence visual descriptions, challenging text scene graph parsers built for single-sentence caption-to-graph mapping. Current approaches typically merge sentence-level parsing outputs for discourse input, often missing phenomena like cross-sentence coreference, resulting in fragmented graphs and degraded downstream VLM task performance. We introduce a new task, \textbf{Disco}urse-level text \textbf{S}cene \textbf{G}raph parsing (\textbf{\DiscoSG{}}), and release \textbf{\DiscoSGDS{}}, a dataset of 400 expert-annotated and 8,430 synthesised multi-sentence caption-graph pairs. Each caption averages 9 sentences, and each graph contains at least 3$\times$ more triples than those in existing datasets. 

Fine-tuning \GPTFour{} on \DiscoSGDS{} yields over 40\% higher SPICE metric than the best sentence-merging baseline. However, its high inference cost and licensing restrict open-source use. Smaller fine-tuned open-source models (e.g., \FlanT) perform well on simpler graphs yet degrade on denser, more complex graphs. To bridge this gap, we introduce \textbf{\DiscoSGRefiner{}}, a lightweight open-source parser that drafts a seed graph and iteratively refines it with a novel learned graph-editing model, achieving 30\% higher SPICE than the baseline while delivering $86{\times}$ faster inference than \GPTFour{}. It generalises from simple to dense graphs, thereby consistently improving downstream VLM tasks, including discourse-level caption evaluation and hallucination detection, outperforming alternative open-source parsers. Code and data are available at~\url{https://github.com/ShaoqLin/DiscoSG}.

\end{abstract}

\section{Introduction}
\label{sec:intro}
\begin{table}[ht!]
    \centering
    %\tiny % Set font size to smallest available
    \renewcommand{\arraystretch}{1.2} % Adjust row height
    \setlength{\tabcolsep}{2pt} % Reduce column spacing
    \begin{tabularx}{\columnwidth}{l X}
        % \toprule
        % 添加图片行
        \footnotesize Image. & \includegraphics[width=0.75\columnwidth]{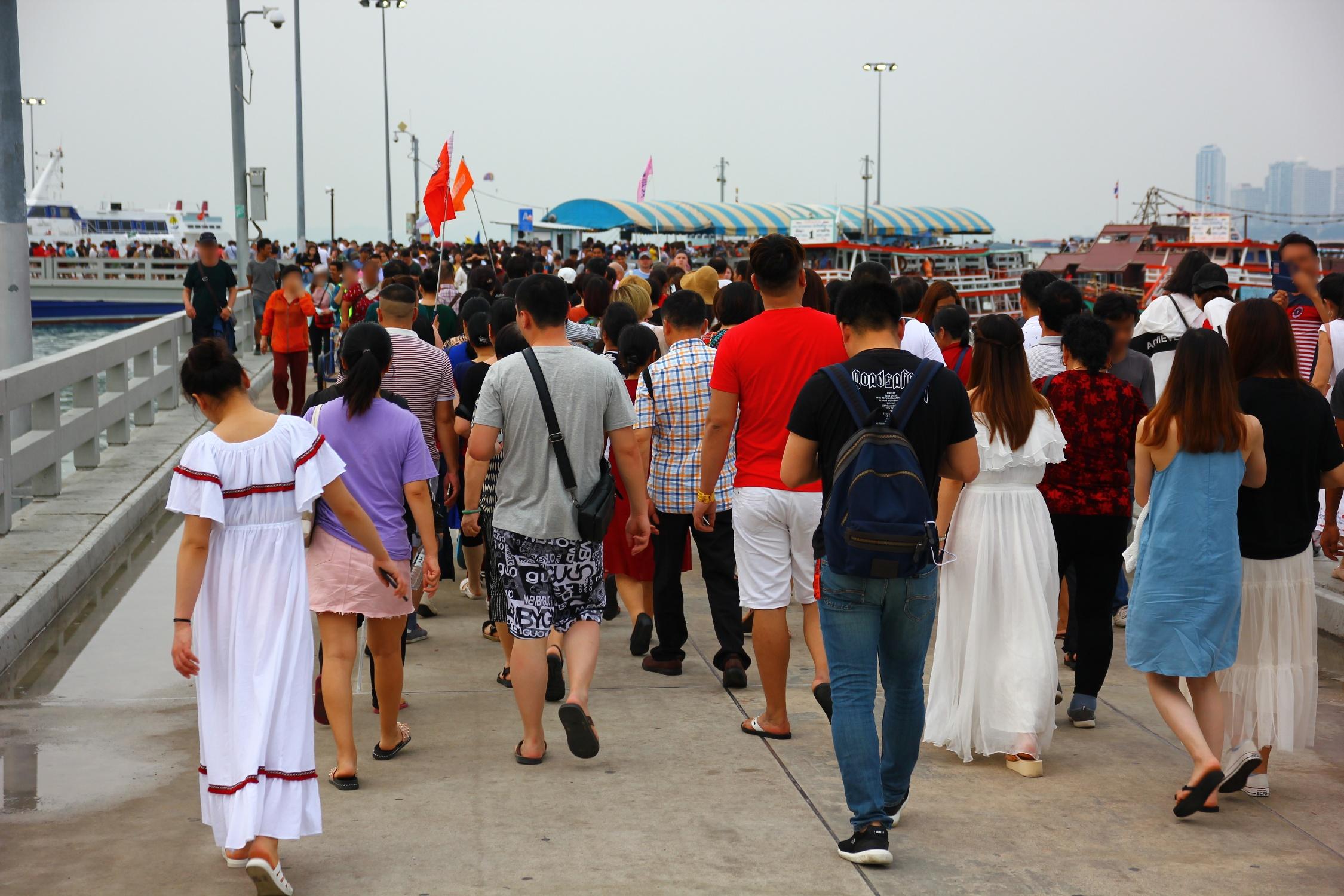} \\
        \hline
        \footnotesize Caption. & \footnotesize A group of people are seen walking on a concrete pier towards a ferry terminal \ldots In the distance, tall buildings loom, indicating that the location is near a city \ldots (\textit{details omitted for brevity})\\
        \hline
        \footnotesize Init.  & \footnotesize (people, walk on, pier), (people, walk towards, ferry terminal), (people, move towards, destination), (pier, is, concrete), (buildings, is, tall) \\
        %\hline
        %\small 6  & \small A \vehiclecolor{truck} driver was \finecolor{fined} after a \loadcolor{large rock} he was transporting fell off at a bend and damaged a passing car worth \$50,000. \\
        \hline
        \footnotesize \textbf{\color{red}Deletion.}  & \footnotesize (people, walk on, pier), (people, walk towards, ferry terminal), \textbf{\color{red}\st{(people, walk towards, destination)}},
        (pier, is, concrete), (buildings, is, tall) \\
        \arrayrulecolor{black!30}\hline\arrayrulecolor{black}
        \footnotesize \textbf{\color{darkgreen}Insertion.}  & \footnotesize (people, walk on, pier), (people, walk towards, ferry terminal), \textbf{\color{red}\st{(people, walk towards, destination)}}, (pier, is, concrete), (buildings, is, tall), \textbf{\color{darkgreen}(people, is, group of)} \\
        \hline
        % & \footnotesize \textbf{Iterative Refinement...} \\
        % \hline
        \footnotesize Refined. & \footnotesize (people, walk on, pier), (people, walk towards, ferry terminal), (pier, is, concrete), (buildings, is, tall), (people, is, group of) \\
        \bottomrule
    \end{tabularx}
    \vspace{-2mm}
    \caption{Illustration of the iterative scene graph refinement process in the \DiscoSGRefiner{} framework.} % , such as \vehiclecolor{Vehicle type}, \loadcolor{Load/Rock description},  \finecolor{Financial action}, \newdetailcolor{Emergent details}.
    \label{tab_eg}
    \vspace{-4mm}
\end{table}
Text scene graph parsing converts visual descriptions into graphs of entities and their relations, supporting tasks like image captioning evaluation~\cite{dong2024benchmarking}, hallucination detection~\cite{yu2024hallucidoctor}, and image retrieval~\cite{johnson2015image}. Traditional methods target single-sentence captions, converting dependency parses into graphs or fine-tuning Pre-trained Language Models (PLMs) on sentence-level pairs. As Vision-Language Models (VLMs) now generate detailed discourse-level, multi-sentence captions with complex inter-sentential dependencies~\cite{cheng2025caparena}, these methods are increasingly inadequate.

Processing discourse-level text introduces several critical challenges for current sentence-level parsers: \textbf{First}, cross-sentence coreference resolution requires correctly linking entities across sentences (e.g., ``a woman'' later as ``she''). \textbf{Second}, capturing long-range relations between entities in different sentences. \textbf{Third}, inferring implicit relationships not stated in any single sentence (e.g., \texttt{(cat, near, window)} from ``The cat is on the mat'' and ``The mat is under the window''). \textbf{Finally}, ensuring graph coherence by producing a globally consistent and complete scene representation. However, existing methods~\cite{dong2024benchmarking} typically resort to heuristically merging sentence-level parsing outputs, resulting in semantically inconsistent graphs that ignore long-range dependencies.

To address these issues, we define \textbf{Disco}urse-level text \textbf{S}cene \textbf{G}raph parsing (\textbf{\DiscoSG{}}), which converts multi-sentence descriptions into scene graphs. We release \textbf{\DiscoSGDS{}}, a dataset of 400 expert-annotated and 8,430 synthesised caption-graph pairs. Each caption averages 9 sentences, and each graph contains at least 3$\times$ more subject-predicate-object triples than prior datasets~\cite{li2023factual,krishna2017visual,yang2025llm}. Our annotation guidelines ensure graphs capture cross-sentence coreference, long-range dependencies, implicit relations, and global consistency.

%\input{tables/discourse_task_evaluation}

% However, standard approaches, such as end-to-end fine-tuning of PLMs~\cite{li2023factual} on caption-graph pairs, face challenges with \DiscoSGDS{}. Fine-tuning large, high-capacity PLMs (e.g., \GPTFour{}\footnote{\url{https://openai.com/index/hello-gpt-4o/}}) perform well, outperforms the state-of-the-art (SOTA) sentence-merging baseline by more than \textbf{40\%} on SPICE metric~\cite{anderson2016spice} across the test split, it is less suitable for open-source use due to slow inference, neither resource-efficient, as a single benchmark run can cost \textbf{approximately \$25 to \$150} in API fees, nor open-source due to licensing constraints. Few-shot prompting of large PLMs~\cite{yang2025llm} also underperforms their fine-tuned counterparts on this discourse task due to the lack of task-specific adaptation. Naive fine-tuning of smaller PLMs (e.g \FlanT{}) performs well, comparable to \GPTFour{}, on the test split with simpler graphs, yet degrades on the test split with denser, more complex graphs, dropping by at least 15\%. and significantly fall behind the sentence-level baseline on downstream VLM tasks that incldues broeader type of graphs, including evaluation of VLM-generated captions and hallucination detection.

However, standard approaches, such as end-to-end fine-tuning of PLMs~\cite{li2023factual} on caption-graph pairs, face challenges with \DiscoSGDS{}. Fine-tuning large PLMs (e.g., \GPTFour{}\footnote{\url{https://openai.com/index/hello-gpt-4o/}}) yields \textbf{over 40\%} higher SPICE~\cite{anderson2016spice} than the strongest sentence-merging baseline. However, it is impractical for open-source use due to slow inference (about \textbf{30s per query}), approximately \textbf{\$25 to \$150} per run in API cost on VLM benchmarks~\cite{dong2024benchmarking}, and restrictive licensing. Few-shot prompting of large PLMs~\cite{yang2025llm} also underperforms their fine-tuned counterparts on this discourse task because it lacks task-specific adaptation. Naive fine-tuning of smaller PLMs (e.g., \FlanT{}~\cite{chung2024scaling}) performs comparably to \GPTFour{} on the simple-graph test set, yet degrades on the dense-graph set by at least \textbf{15\%} SPICE, and falls significantly behind the sentence-merging baseline on downstream VLM tasks that involve a broad range of graph complexities, including evaluation of \textbf{VLM-generated captions and hallucination detection}.

% To overcome these limitations, we propose \textbf{\DiscoSGRefiner{}}, a lightweight, iterative graph-refinement framework (Table~\ref{tab_eg}). It starts from a base graph obtained by merging sentence-level parses, then iteratively improves graph quality through edit operations proposed by an encoder-decoder PLM-based ``Programmer''. This module \textbf{innovatively} disentangles edit operations: the encoder identifies and removes erroneous and inconsistent triples, while the decoder generates new triples to enhance completeness and semantic coverage. The Programmer is trained on \DiscoSGED{}, a synthetic corpus of edit operations derived from \DiscoSGDS{}. Operating in edit space substantially reduces the complexity of generating dense, complex graphs while incorporating broader inter-sentence context and task-specific knowledge. Built from two \FlanTBase{} models~\cite{chung2024scaling} totalling \textbf{0.5B} parameters, \DiscoSGRefiner{} achieves \textbf{30\%} higher SPICE than the sentence-merging baseline, delivers \textbf{86$\times$} faster inference than \GPTFour{}, and runs on a single GPU at substantially lower cost. It maintains strong performance on both simple- and dense-graph subsets and \textbf{consistently improves discourse-level caption evaluation and hallucination detection}, outperforming the strongest sentence-merging parser and end-to-end fine-tuned \FlanT{} parsers.

To overcome the limitations, we propose \textbf{\DiscoSGRefiner{}}, a lightweight iterative graph-refinement framework (Table~\ref{tab_eg}). Starting from a base graph merged from sentence-level parses, an encoder-decoder PLM ``Programmer'' proposes edits to simplify dense graph construction. We \textit{disentangle} deletion and insertion so the decoder generates shorter edit sequences, which improves editing performance: the encoder deletes erroneous triples, and the decoder inserts triples to improve semantic coverage. Trained on \DiscoSGED{}, a synthetic corpus of edit operations derived from \DiscoSGDS{}, the Programmer learns to exploit the inter-sentence context and task knowledge in \DiscoSGDS{}.
Built from two \FlanTBase{} models totalling \textbf{0.5B} parameters, \DiscoSGRefiner{} \textbf{balances generalisation, performance, and cost}. It achieves \textbf{30\%} higher SPICE than the sentence-merging baseline on both simple and dense graph test sets, \textbf{without degradation} across graph densities; delivers \textbf{86$\times$} faster inference than \GPTFour{}; and runs on a single GPU at substantially lower cost. On downstream VLM tasks, it consistently improves results, outperforming the strongest sentence-merging parser and end-to-end \FlanT{} parsers.

Due to the lack of resources for discourse-level hallucination detection in VLMs, we introduce \DFOIL{}, a benchmark inspired by \FOIL{}~\cite{shekhar2017foil}, originally designed for sentence-level detection. \DFOIL{} comprises \textbf{200} pairs of hallucinated and human-corrected \textbf{multi-sentence} captions, with corrections made against a reference.

\textbf{Overall, our main contributions are:} \textbf{\textsc{(i)}} defining \DiscoSG{}, a novel task for discourse-level text scene graph parsing from multi-sentence visual descriptions; \textbf{\textsc{(ii)}} introducing \DiscoSGDS{}, a dataset of 400 expert-annotated and 8,430 synthesised examples for discourse-level parsing; \textbf{\textsc{(iii)}} proposing \DiscoSGRefiner{}, a lightweight iterative framework with a novel PLM-based ``Programmer'' trained on synthesised edits (\DiscoSGED{}) derived from \DiscoSGDS{}; \textbf{\textsc{(iv)}} demonstrating that our 0.5B open-source model outperforms strongest sentence-merging baselines by 30\%, and is more resource-efficient than fine-tuned \GPTFour{}; and \textbf{\textsc{(v)}} establishing \DFOIL{}, a new benchmark for discourse-level hallucination detection.

%a lightweight iterative framework with a PLM-based “Programmer” trained on edits (\DiscoSGED{}) derived from \DiscoSGDS{};

%\section{\DiscoSG: Task Definition}
%\label{sec:task_definition}
%\input{02_task}

\section{DiscoSG-DS: Discourse-Level Scene Graph Dataset}
% Overview paragraph (Keep your concise version)
To study \DiscoSG{}, we introduce the \DiscoSGDS{} dataset, which consists of 400 manually annotated gold-standard examples and 8,430 synthesised instances. Each instance aligns a discourse-level image description $\vx = (x_1, \dots, x_n)$ with a corresponding scene graph $\vy$. Each description $\vx$ contains $n$ sentences that are contextually dependent, requiring parsers to model discourse-level semantics. Each graph ($\vy$) captures the semantics of $\vx$ through a set of structured triples representing object relationships $(e_{sub}, r, e_{obj})$ (such as \texttt{(man, wear, hat)}) and attributes $(e, a, v)$ (like \texttt{(hat, has\_attribute, red)}).

%This includes 400 manually annotated gold-standard examples and 8,430 synthesised instances generated via a fine-tuned GPT-4o model. Unlike sentence-level parsing, which typically targets short captions, \DiscoSG{} handles discourse-level image descriptions $\vx = (x_1, \dots, x_n)$ consisting of multiple, contextually dependent sentences, requiring parsers with discourse-level understanding. Each graph $\vy$ captures the semantics of $\vx$ through a set of structured triples, including object relationships $(e_{\text{sub}}, r, e_{\text{obj}})$ (e.g., \texttt{(man, wear, hat)}) and attributes $(e, a, v)$ (e.g., \texttt{(hat, has\_attribute, red)}).

% It includes 400 manually annotated gold-standard examples and 8,430 synthesised instances generated by a fine-tuned GPT-4o model.

% Sub-subsection: Where the initial text came from
\subsection{Source Data Selection}
\label{ssubsec:source_selection}
% Descriptions ($\vx$) were sourced from the \SharedGPTV{} dataset \cite{chen2024sharegpt4v}, containing 1.2 million captions generated by GPT-4 for images. These highly descriptive captions generally span multiple sentences, making them suitable for discourse-level analysis. % from \textsc{LCS}~\cite{li2022blip} and \textsc{MSCOCO}~\cite{vinyals2016show} The high quality of these synthetic descriptions, comparable to human annotations, was also validated in the original SharedGPT4V work. 
% We selected 40,000 diverse image-description pairs using diversity sampling based on TF-IDF text embeddings, adapting methods from~\citet{zhuo2023robustness,li2023best}. The selected descriptions were then annotated either manually or via synthesis to form the final \DiscoSGDS{} dataset.
Descriptions ($\vx$) were sourced from the \SharedGPTV{} dataset~\cite{chen2024sharegpt4v}, containing 1.2 million captions generated by GPT-4 for images. These highly descriptive captions span multiple sentences, making them suitable for discourse-level analysis. We selected 40,000 diverse image-description pairs using diversity sampling based on TF-IDF text embeddings, adapting methods from~\citet{zhuo2023robustness,li2023best}. The selected descriptions were then annotated either manually or via synthesis to form the final \DiscoSGDS{} dataset.

% Sub-subsection: How the 400 manual examples were made
\subsection{Manual Annotation Pipeline}
\label{subsec:manual_pipeline}
To construct high-quality ground truth data for \DiscoSG{}, we manually annotated 400 descriptions randomly selected from the pool of 40,000, using a human-in-the-loop active learning framework. 

\paragraph{Annotator Setup and Annotation Principles.}
Annotation was conducted by two expert annotators: a senior researcher with over three years of relevant task experience and a trained postgraduate student in computer science. To calibrate the postgraduate annotator, we conducted a phased training process in which the student annotated 40 examples across four batches under the postdoc’s supervision. By the final batch, the annotators had achieved 90\% inter-annotator agreement. %These calibration examples were excluded from the final dataset.

Text descriptions were first annotated using the FACTUAL-MR intermediate representation~\cite{li2023factual}, which reduces annotation ambiguity and enforces a standardised structure, and were then converted into formal scene graphs. The annotation focused on resolving cross-sentence coreference, correcting quantifier inconsistencies, capturing missing long-range relational dependencies, and identifying overlooked implicit relations and semantic contradictions. To ensure the resulting scene graphs accurately reflected the visual content, the protocol also required annotators to correct factual inconsistencies or hallucinations found in the source descriptions ($\vx$) by verifying against the corresponding images. Detailed annotation guidelines are provided in~\Cref{app:guidelines}.

\paragraph{Initial Set Creation.}
We initiated the human-in-the-loop process by generating draft scene graphs for 102 image descriptions, created through heuristic merging of sentence-level parses from FACTUAL-T5~\cite{li2023factual}. Each draft underwent two-stage refinement: the postgraduate annotator first edited the graph, followed by review and finalisation by the postdoctoral researcher, with consensus required for acceptance. From this curated set, 40 examples were reserved as a fixed validation set ($D_{val}$), and the remaining 62 formed the seed training set ($D_{seed}$) for active learning. 

% \input{tables/tab-algo1} % Algorithm associated with Active Learning
% \paragraph{Active Learning.} % Step 3
% We expanded the manual training data through two iterations of active learning (\Cref{alg:active-learning} at \Cref{app:al_algo}). Starting with GPT‑4o model $M_0$ fine-tuned on $D_{seed}$, each iteration involved: (i) generating draft scene graphs for a new batch of unlabeled data $B_{raw}$, (ii) refining these drafts through the same two-stage expert review to produce $B_{refined}$, (iii) adding $B_{refined}$ to the growing training set $D_{train}$, and (iv) fine-tuning GPT‑4o again to obtain the next model $M_{i+1}$.
% $D_{train}$ grew from 62 to 156 and then to 360 instances. This process significantly improved GPT‑4o's performance on $D_{val}$ (SPICE score~\cite{anderson2016spice}: XX $\rightarrow$ YY $\rightarrow$ ZZ) and reduced the postgraduate annotator’s average annotation time from approximately 30 minutes to 10 minutes per instance. Finally, we merged $D_{train}$ and $D_{val}$ and re-split the 400 examples into 300 training and 100 test instances.

\paragraph{Active Learning.}
% We adopt an iterative active learning loop that can, in principle, run until performance saturates.  
% In practice, validation gains plateaued and our annotation budget was fixed, so we carried out \textbf{two} iterations (pseudocode in \Cref{alg:active-learning} at Appendix~\ref{app:al_algo}).  
% Starting from GPT‑4o model $M_0$ fine‑tuned on the 62‑example seed set $D_{\text{seed}}$, each iteration proceeded as follows:
We adopted an iterative active learning loop and carried out \textbf{two} iterations, as validation gains plateaued and our annotation budget was fixed (pseudocode in \Cref{alg:active-learning}, Appendix~\ref{app:al_algo}). Starting from the initial GPT‑4o model $M_0$, fine‑tuned on the 62-example seed set $D_{\text{seed}}$, each iteration proceeded as follows:

\textsc{i)} \textbf{Batch selection}: We randomly sampled descriptions, yielding batches $B_{\text{raw}}$ of 94 instances in round 1 and 204 in round 2.

\textsc{ii)} \textbf{Draft generation}: The current model $M_i$ produced a scene graph draft for every selected description in the batch.

\textsc{iii)}  \textbf{Two‑stage review}: The student annotator corrected each draft and the postdoc validated it, yielding refined graphs $B_{\text{refined}}$, which were then added to the training set ($D_{\text{train}} = D_{\text{train}} \cup B_{\text{refined}}$).
 
\textsc{iv)}  \textbf{Model update}: GPT‑4o was fine‑tuned on the updated training set $D_{\text{train}}$ to obtain the next model $M_{i+1}$.

After two rounds, the training set size $|D_{\text{train}}|$ increased from 62 to 360 examples while the annotation time per instance fell from $\sim$30 min to 10 min. Correspondingly, \GPTFour{}’s SPICE on the validation set $D_{\text{val}}$ increased from 70.8 to 73.3 to 76.1.
We merged $D_{\text{train}}$ and $D_{\text{val}}$ and created two 300/100 train-test splits of the 400 human-annotated graphs.

%SPICE scores~\cite{anderson2016spice} on the validation set ($D_{\text{val}}$, 40 examples) rose from \textsc{XX}$\!\rightarrow\!$\textsc{YY}$\!\rightarrow\!$\textsc{ZZ}, while

% Sub-subsection: How the 8430 synthesised examples were made
\subsection{Synthesis Annotation Pipeline}
\label{ssubsec:synthesis}
To further scale the data, we first used GPT‑4o to filter out descriptions that showed hallucination errors when compared with their images, retaining 8,430 high-quality descriptions. We then used the teacher model, a GPT‑4o fine-tuned on 300 human-annotated training examples, to generate scene graph annotations for the remaining descriptions. The teacher model achieves a high SPICE score (F1 triple matching) of 73\% on the test set.
\begin{table}[t]
  \centering
  \resizebox{\columnwidth}{!}{%
  % Added vertical lines "|" to the tabular specification
  % Using @{\hspace{<len>}} to reduce space between columns (e.g., 4pt)
  % Note: Vertical lines are generally discouraged with booktabs rules
  \begin{tabular}{|l@{\hspace{4pt}}|r@{\hspace{4pt}}|r@{\hspace{4pt}}|r@{\hspace{4pt}}|r@{\hspace{4pt}}|r@{\hspace{4pt}}|r|} % Added | separators
\hline
    % Shorter abbreviations, corrected Avg Len
    \textbf{Dataset} & \textbf{\# Inst.} & \textbf{Avg Len} & \textbf{Avg Trp} & \textbf{Avg Obj} & \textbf{Avg Rel} & \textbf{Total Trp} \\
    \midrule
    % Removed citations from VG and FACTUAL rows
    \VG{} & 2,966,195 & 5.34 & 1.53 & 1.69 & 1.22 & 4,533,271 \\
    \FACTUAL{} & 40,369 & 6.08 & 1.76 & 2.12 & 1.57 & 71,124 \\
    \TSGBench{} & 2,034 & 12.23 & 5.81 & 5.63 & 3.65 & 11,820 \\
\hline
    % Combined Human row
    \DiscoSGDS{}  &  &  &  &  &  &  \\
    \ \ \ \ Human & 400 & 181.15 & 20.49 & 10.11 & 6.54 & 8,195 \\
    \ \ \ \ Synthetic & 8,430 & 163.07 & 19.41 & 10.06 & 6.39 & 163,640 \\
\hline
  \end{tabular}%
  }
  % Removed duplicate caption
  \vspace{-3mm}
\caption{Comparison of dataset statistics. Columns detail the number of instances (\# Inst.), average description length in words (Avg Len), average counts per instance for triples (Avg Trp), unique objects (Avg Obj), and unique relations (Avg Rel), along with the total number of triples (Total Trp).}
\label{tab:dataset_statistics}
\vspace{-5mm}
\end{table}

\subsection{Dataset Statistics and Analysis}
\label{subsec:stats_analysis}
The resulting \DiscoSGDS{} dataset comprises 300 human-annotated training and 100 test instances, along with 8,430 synthesised examples. Each instance includes an average of 9.3 sentences. \Cref{tab:dataset_statistics} compares \DiscoSGDS{} with other datasets that contain \textbf{only single-sentence} captions, including Visual Genome (\VG)~\cite{krishna2017visual}, \FACTUAL{}~\cite{li2023factual}, and \TSGBench{}~\cite{yang2025llm}. On average, each \DiscoSGDS{} graph contains roughly 15$\times$ more triples than those in \VG{} or \FACTUAL{} and 3$\times$ than those in \TSGBench{}, and demonstrates greater lexical diversity across longer spans of caption text (see Appendix~\ref{sub:lexical}), highlighting its discourse-level complexity.

\section{DiscoSG-Refiner: Iterative Graph Refinement for Scene Graph Parsing} 
% Our goal is to develop an effective and computationally efficient discourse-level parser $P(\vy|\vx;\vtheta)$. Traditional approaches, such as FACTUAL-T5, adapt PLMs for this task, but direct generation of flattened graph sequences faces several challenges:

% \textsc{i)} \textbf{Fine-tuning smaller PLMs}, particularly those below 1.5B parameters (e.g., versions of Flan-T5~\cite{chung2024scaling}, Qwen~\cite{bai2023qwen}, and Llama~\cite{grattafiori2024llama}), struggle with the required generation length and long-range dependencies in discourse inputs, often producing repetitive graphs or outputs with syntactic or semantic errors, consistent with known limitations in PLM long sequence generation~\cite{xu2022learning}.

% \textsc{ii)} \textbf{Fine-tuning large-scale PLMs} (comparable to GPT-4o) can produce higher-quality scene graph outputs but remains incompatible with our open-source goals due to prohibitive computational costs, deployment challenges, and restrictions from closed-source licensing.

% \textsc{iii)} \textbf{Few-shot prompting of large PLMs like GPT-4o}~\cite{zhang2024path} also underperforms on this structured prediction task, likely due to insufficient domain and task adaptation.

%Our goal is to develop an effective and computationally efficient discourse-level parser $P(\vy|\vx;\vtheta)$. 
Formally, given a multi-sentence image description $\vx$, the goal of \DiscoSG{} is to learn a parameterised model that predicts the most probable scene graph $\vy' = \operatorname*{argmax}_{\vy \in \mathcal{Y}} P(\vy \mid \vx; \vtheta)$, where $\mathcal{Y}$ is the set of all possible scene graphs. Existing approaches include fine-tuning small PLMs~\cite{li2023factual,sharifzadeh2022improving}, few-shot prompting with large PLMs~\cite{yang2025llm}, or fine-tuning large PLMs such as our teacher model. However, these methods either perform poorly on discourse-level inputs or incur high computational costs. The approach of merging sentence-level parser outputs~\cite{dong2024benchmarking} offers a better performance-efficiency trade-off but still fails to capture inter-sentence dependencies.

% \textsc{i)} \textbf{Fine-tuning smaller PLMs} (e.g., low-parameter variants of Flan-T5~\cite{chung2024scaling}, Qwen~\cite{bai2023qwen}, and Llama~\cite{grattafiori2024llama}) often fails due to limited capacity to handle long-range dependencies and produce long outputs, often resulting in repetitive content or syntactically invalid graphs~\cite{xu2022learning}. End-to-end methods like fine-tuning small PLMs (e.g., FACTUAL-T5), fine-tuning large PLMs (e.g., GPT-4o), or few-shot prompting of large PLMs~\cite{zhang2024path} to generate flattened graph sequences all encounter their respective challenges with discourse-level inputs.

% \textsc{ii)} \textbf{Fine-tuning large PLMs} (e.g., GPT-4o-scale models) yields better quality but is impractical for open-source deployment due to high computational costs and licensing constraints.

% \textsc{iii)} \textbf{Few-shot prompting of large PLMs} (e.g., GPT-4o~\cite{zhang2024path}) performs poorly on this structured prediction task, likely due to insufficient task-specific adaptation.

We propose \DiscoSGRefiner{}, a novel iterative refinement method using small open-source PLMs that mirrors our human annotation workflow. Inspired by machine translation post-editing~\cite{vu2018automatic,li2024improving} and text-to-graph generation~\cite{han2024pive}, it first produces a flawed initial parse and then applies targeted edits to yield the final scene graph, dramatically reducing the required generation overhead.

\subsection{Framework Overview}
\label{subsec:framework}

\DiscoSGRefiner{} extends the Generator-Programmer-Interpreter framework~\cite{vu2018automatic,li2024improving} with task-specific adaptations across its three modules, placing particular focus on a novel Programmer optimised for refining long discourse-level scene graphs. The process begins with an initial graph $\vyO$ generated by the Generator. Then, over T steps, the Programmer proposes actions $\va^t$ which a deterministic Interpreter applies to produce $\vytnext$, iterating until the final graph $\vy = \vyT$ is obtained.

Formally, the probability of generating the final graph $\vy = \vyT$ given the input $\vx$ is defined as:

{\small
\setlength{\abovedisplayskip}{3pt}
\setlength{\belowdisplayskip}{6pt}
\begin{equation}
\label{eq:iterative_prob_with_interp}
\begin{aligned}
P(\vy^{T}|\vx;\vtheta) &= P_{\mathrm{Gen}}(\vy^{0}|\vx) \\
\times \prod_{t=0}^{T-1} & \bigl(P_{\mathrm{Prog}}(\va^{t}|\vy^{t},\vx;\vtheta) \cdot P_{\mathrm{Intp}}(\vy^{t+1}|\vy^{t},\va^{t})\bigr)
\end{aligned}
\end{equation}
}
% {\setlength{\abovedisplayskip}{3pt}
% \setlength{\belowdisplayskip}{5pt}
% \begin{equation}
% \label{eq:iterative_prob_with_interp}
% \small
% \begin{aligned}
% P(\vy^{T}|\vx;\vtheta) &= P_{\mathrm{Gen}}(\vy^{0}|\vx) \\
% \times \prod_{t=0}^{T-1} & \bigl(P_{\mathrm{Prog}}(\va^{t}|\vy^{t},\vx;\vtheta) \cdot P_{\mathrm{Intp}}(\vy^{t+1}|\vy^{t},\va^{t})\bigr)
% \end{aligned}
% \end{equation}
% }
\subsubsection{Generator}
\label{subsubsec:generator}
The initial graph $\vyO$ probability is derived from applying a sentence-level parser $P_{\text{Sent}}$ independently to each sentence $x_i \in \vx$, such that $P_{Gen}(\vyO|\vx) \propto \prod_{i=1}^{n} P_{\text{Sent}}(y_{i} | x_{i})$. Operationally, we generate $\vyO$ using \FACTUALT as $P_{\text{Sent}}$ to parse each $x_i$ into sentence-level graphs $\{y_1, \dots, y_n\}$, which are then merged (treating nodes with identical names as co-referent) to form the initial graph. This provides broad coverage of explicit mentions but often introduces errors of commission (redundancies) and omission (missing cross-sentence or implicit relations), requiring subsequent refinement.

\subsubsection{Programmer}
\label{subsubsec:programmer}
At each step $t$, Programmer ($P_{\mathrm{Prog}}(\va^{t}|\vy^{t},\vx;\vtheta)$), our core learnable component with parameters $\vtheta$, generates a set of edit actions $\va^t = (\mathbf{D}^t, \mathbf{I}^t)$, which includes deletion $\mathbf{D}^t$ and insertion $\mathbf{I}^t$ operations based on the current graph $\vyt$ and the input $\vx$. 

Directly applying previous post-editing techniques is suboptimal. Methods generating combined deletion/insertion sequences~\cite{li2024improving} require overly long generation for small PLM decoders, while insert-only methods~\cite{han2024pive}, although having shorter generation length, cannot correct the errors of commission present in our initial $\vy^0$. We observe that approximately 1/3 of the initial graph triples require deletion. %requiring an approach that supports both operations effectively.

Therefore, we propose a \textbf{novel} Programmer architecture using an encoder-decoder PLM (e.g., \FlanT{}) that \textit{disentangles deletion and insertion prediction to reduce decoder generation length}:

% \paragraph{Deletion Prediction (Encoder-based):} The encoder takes $\vx$ and $\vy^t$ as input. The objective is to identify which triples to delete, where the triple representation is from average contextual representations within the two brackets of the triple obtained from encoder. Specifically, the encoder predicts \texttt{KEEP[$t_i$]} (0) or \texttt{DELETE[$t_i$]} (1) actions for each triple $t_i$ in the flattened representation of $\vy^t$.

\noindent\textbf{Deletion Prediction (Encoder-based):} The encoder takes $\vx$ and $\vy^t$ as input. The objective is to identify which triples to delete, with each triple represented by the average of contextual representations of tokens between its opening and closing brackets. Specifically, the encoder predicts \texttt{KEEP[$t_i$]} (0) or \texttt{DELETE[$t_i$]} (1) action flags through a binary classification layer for each triple $t_i$ in the flattened representation of $\vy^t$.

\noindent\textbf{Insertion Generation (Decoder-based):} The decoder, conditioned on the encoder's representation of $\vx$ and $\vy^t$, generates token sequences forming \textit{unseen graph triples}, each generated token $t_j$ implicitly representing an \texttt{INSERT[$t_j$]} action. This approach significantly reduces generation length compared to regenerating the entire graph.

% or outputs \texttt{[No\_Action]} if no insertions are needed

\subsubsection{Interpreter}
\label{subsubsec:interpreter}
% The deterministic Interpreter applies the proposed action $\mathbf{a}^t = (\mathbf{D}^t, \mathbf{I}^t)$ to transform the current graph $\vyt$ into the next state $\vytnext$. It handles both deletion of existing triples and insertion of new ones based on the Programmer's predictions. This corresponds to the term $P_{\mathrm{Intp}}(\vy^{t+1}|\vy^{t},\mathbf{a}^{t})$ in our equation, which is deterministic since the next graph state is fully determined by the edit actions applied to the current state.
%Although the Programmer generates token-level outputs ($D^t$ contains token classifications, $I^t$ is a token sequence) to leverage the PLM's token-based architecture, the Interpreter translates these into triple-level graph modifications.

The Interpreter ($P_{\mathrm{Intp}}(\vy^{t+1}|\vy^{t},\va^{t})$) applies the edit actions $\va^t = (\mathbf{D}^t, \mathbf{I}^t)$ to transform $\vyt$ into $\vytnext$. Because this process is deterministic, the probability $P_{\mathrm{Intp}}(\vytnext | \vyt, \va^t)$ is simply 1 for the unique resulting state $\vytnext$ (and 0 otherwise). The Interpreter operates in two stages:

%\paragraph{For deletions}, $\mathbf{D}^t$ contains token-level binary predictions. The Interpreter removes any triple where at least one token is marked for deletion.

\noindent\textbf{Deletion:} With deletion flags $\mathbf{D}^t$, the Interpreter removes the identified triples $\mathcal{D}^t_{\text{triples}}$ from $\vyt$.

%\paragraph{For insertions}, $\mathbf{I}^t$ consists of generated token sequences; The token sequences are then parsed into valid triples, with grammatically incorrect generations filtered out.  into well-formed triples and filters out any grammatically incorrect constructions. 

\noindent\textbf{Insertion:}  
The token sequence $\mathbf{I}^t$ generated by the decoder is parsed into candidate triples. The Interpreter then validates these for structural correctness (e.g., subject-relation-object format) and filters out any malformed or incomplete ones. This yields the set of valid triples $\mathcal{I}^t_{\text{triples}}$, which are subsequently inserted into $\vy^t$ as an unordered collection.

% The resulting graph state $\vy^{t+1}$ is obtained by first deleting  $\mathcal{D}^t_{\text{triples}}$ from $\vy^t$, followed by the insertion of  $\mathcal{I}^t_{\text{triples}}$: $\vytnext = (\vyt \setminus \mathcal{D}^t_{\text{triples}}) \cup \mathcal{I}^t_{\text{triples}}$.

The resulting graph state for the next iteration follows a delete-first-then-insert heuristic: $\vytnext = (\vyt \setminus \mathcal{D}^t_{\text{triples}}) \cup \mathcal{I}^t_{\text{triples}}$ . %This ensures the Programmer's token-level predictions are applied as valid triple-level graph updates.

% The final state for the iteration is then produced by applying these derived triple edits: $\vytnext = (\vyt \setminus D^t_{\text{triples}}) \cup I^t_{\text{triples}}$. This deterministic process ensures the Programmer's token-level predictions are consistently translated into valid graph updates.

% For both encoder and decoder, we translate from token-level operations to structured triples-level operations, which take advantage of the token-based language model capabilities for our triple-based graph manipulation. This approach effectively translates token-level operations into triple-level graph manipulations.

%\subsubsection{Iteration Process}
%\label{subsubsec:iteration}
%This refinement process iterates for a fixed number of iterations $T$ or until convergence, allowing progressive correction of the graph. Our architecture efficiently handles both types of edits while keeping generation length manageable for smaller models and maintaining the conceptual integrity of triple-level graph operations.

\subsection{Training the Programmer}
\label{subsec:training}
%The Programmer is trained via supervised learning with .
\paragraph{Edit-Annotated Data (DiscoSG-ED) Generation.}
Training requires editing annotations derived from \DiscoSGDS{}. For each instance $(\vx, \vy_{gold})$ from both the manual and synthetic sets, we generate the initial graph $\vyO$ given $\vx$ using Generator. We also create alternative versions of $\vy^{(0)}$ by synthetically corrupting $\vy_{\text{gold}}$ via the random deletion or insertion of graph triples. The ground-truth edit actions $(\mathbf{D}_{gt}, \mathbf{I}_{gt})$ are derived by comparing both types of $\vyO$ against $\vy_{gold}$. 

% The ground-truth deletion flags $\mathbf{D}_{gt}$ is derived by finding the set of triples $\mathcal{D}_{gt}$ present in $\vyO$ but not in $\vy_{gold}$. 
% As insertion is token-level, after finding the triples $\mathcal{I}_{gt}$ present in $\vy_{gold}$ but not in $\vyO$, we convert it into token sequences, the ground-truth insertions $\mathbf{I}_{gt}$ for the decoder to generate. These derived tuples $(\vx, \vyO, \mathbf{D}_{gt}, \mathbf{I}_{gt})$ form our training dataset, \DiscoSGED{}. Depending on the specific augmentation hyper-parameters used, the size of \DiscoSGED{} can be 10-20 times that of \DiscoSGDS{}. Detailed dataset statistics are provided in the Appendix.

The ground-truth deletion flags $\mathbf{D}_{gt}$ are derived by finding the set of triples $\mathcal{D}_{gt}$ present in $\vyO$ but not in $\vy_{gold}$. 
For insertion, after identifying triples $\mathcal{I}_{gt}$ present in $\vy_{gold}$ but not in $\vyO$, we convert triples into token sequences to form the ground-truth insertions $\mathbf{I}_{gt}$ to train the decoder. These derived tuples $(\vx, \vyO, \mathbf{D}_{gt}, \mathbf{I}_{gt})$ form our training dataset, \DiscoSGED{}. If we generate $N$ corrupted versions per gold graph, the size of \DiscoSGED{} becomes $N$ times that of \DiscoSGDS{}.
%Detailed dataset statistics are provided in the Appendix.

%To adapt this triple-level comparison to our token-based implementation, each token $t_i$ within the flattened representation of $\vyO$ receives a binary label (1 if $t_i \in \mathcal{D}_{gt}$, 0 otherwise). 

\paragraph{Loss Functions and Optimisation.}
The Programmer's encoder and decoder are jointly trained via multitask learning, optimising separate supervised objectives. The encoder is trained using a binary cross-entropy loss ($\mathcal{L}_\text{delete}$). The decoder is trained using a standard sequence-to-sequence cross-entropy loss ($\mathcal{L}_\text{insert}$) to generate the target insertion sequence $\mathbf{I}_{gt}$. The final loss is a weighted sum ($\mathcal{L} = \lambda_\text{del}\mathcal{L}_\text{delete} + \lambda_\text{ins}\mathcal{L}_\text{insert}$).

\section{Experiments}
We evaluate \DiscoSGRefiner{} through three complementary analyses: (i) benchmark performance on the \DiscoSGDS{} test set, (ii) impact on downstream parsing tasks, and (iii) ablation studies to isolate the contribution of each component.

\subsection{Discourse-Level Text Scene Graph Parsing Evaluation}

\paragraph{Dataset.} We evaluate on the \DiscoSGDS{} \textbf{Random} and \textbf{Length} test sets (100 each). \textit{Random} is a uniform split, while \textit{Length} sorts by the number of graph triples, so training covers fewer-triple graphs and testing targets more complex graphs, assessing generalisation to long, complex graphs.

\paragraph{Metrics.} Parser performance is measured by comparing the predicted scene graph against the gold graph using these metrics: 1) \textbf{SPICE}~\cite{anderson2016spice}, an F1 score for exact-match triple overlap; 2) \textbf{Bi-SoftSPICE (BSSPICE)}, a symmetrized version of SoftSPICE~\cite{li2023factual} using semantic similarities between triple embeddings measuring graph similarities; 3) \textbf{computational efficiency metrics} (i.e., inference time\footnote{For methods using models $\geq$70B parameters and \GPTFour{}, inference time is measured via API call time per query.} and model parameter count); and 4) \textbf{discourse-specific error rate (\%)}, measuring four types of discourse-level parsing errors based on \GPTFour{} and human assessments. For each instance, if an error type is present in the graph, it is counted as one error regardless of how many times it occurs. Detailed error definitions are provided in~\Cref{app:error}.
{
\setlength{\abovedisplayskip}{3pt}
\setlength{\belowdisplayskip}{5pt}
\begin{table*}[t]
\centering
\resizebox{\textwidth}{!}{%
\begin{tabular}{llcccccc}
\toprule
\rowcolor{white}
& & \multicolumn{2}{c}{\textbf{Random}} & \multicolumn{2}{c}{\textbf{Length}} & & \\
\textbf{Category} & \textbf{Method} & \textbf{SPICE}\,$\uparrow$ & \textbf{BSSPICE}\,$\uparrow$ & \textbf{SPICE}\,$\uparrow$ & \textbf{BSSPICE}\,$\uparrow$ & \textbf{Time (s)}\,$\downarrow$ & \textbf{Param.}\,$\downarrow$ \\
\midrule
% \multicolumn{7}{l}{\textbf{\large Sentence Parsing \& Merging}} \\
\multirow{3}{*}{\textbf{\begin{tabular}[c]{@{}l@{}} \large Sentence Parsing \& Merging \end{tabular}}}
&\StanfordParser{}                      & 17.0 & 81.5 & 19.5 & 83.1 & \textasciitilde0.21s  & - \\
&\VGTBaseSentMerge{}         & 45.3 & 89.4 & 45.9 & 90.1 & \textasciitilde0.19s & 0.25B \\
&\FACTUALTBaseSentMerge{}       & 49.4 & 90.9 & 52.7 & 92.0 & \textasciitilde0.19s & 0.25B \\
\midrule
% \multicolumn{7}{l}{\textbf{\large End-to-End (Sentence)}} \\
\multirow{2}{*}{\textbf{\begin{tabular}[c]{@{}l@{}} \large End-to-End (Sentence)\end{tabular}}}
%\arrayrulecolor{black!30}\hline\arrayrulecolor{black}
%\multicolumn{5}{l}{\textbf{VG Training}} \\
&\VGTBaseDirect                         & 15.3 & 82.5 & 13.8 & 82.4 & \textasciitilde0.07s & 0.25B \\
%\arrayrulecolor{black!30}\hline\arrayrulecolor{black}
%\multicolumn{5}{l}{\textbf{VG + FACTUAL Training}} \\
&\FACTUALTDirect{}             & 37.6 & 88.0 & 32.2 & 87.5 & \textasciitilde0.07s & 0.25B \\
\midrule
% \multicolumn{7}{l}{\textbf{\large End-to-End (Discourse)}} \\
% \arrayrulecolor{black!30}\hline\arrayrulecolor{black}
% \multicolumn{7}{l}{\textbf{\DiscoSGDS{} (8730 examples)}} \\
\multirow{6}{*}{\textbf{\begin{tabular}[c]{@{}l@{}}\large End-to-End (Discourse)\\ \textit{\textbf{\DiscoSGDS{} (8730 examples)}}\end{tabular}}}
% \DiscoSGTFiveBase{}-1epoch                      & 49.2 & 90.8 & xx.x & xx.x & \textasciitilde0.07s & 0.25B \\
% \DiscoSGTFiveLarge{}-1epoch                     & 63.6 & 94.0 & xx.x & xx.x & \textasciitilde0.32s & 0.78B \\
% \DiscoSGTFiveXL{}-1epoch                 & xx.x & xx.x & xx.x & xx.x & \textasciitilde0.39s & 3B \\
% \DiscoSGQwenHalfB{}-1epoch                & 41.6 & 87.0 & xx.x & xx.x & \textasciitilde0.39s & 0.5B \\
% \DiscoSGQwenOneFiveB{}-1epoch             & 7.5 & 68.1 & xx.x & xx.x & \textasciitilde0.77s & 1.5B \\
% \DiscoSGQwenSevenB{}-1epoch                 & xx.x & xx.x & xx.x & xx.x & \textasciitilde3.27s & 7B \\
% \midrule
% \DiscoSGTFiveBase{}-1epoch+earlystop                      & 56.7 & 92.4 & xx.x & xx.x & \textasciitilde0.07s & 0.25B \\
% \DiscoSGTFiveLarge{}-1epoch+earlystop                     & 61.2 & 93.4 & xx.x & xx.x & \textasciitilde0.32s & 0.78B \\
% \DiscoSGTFiveXL{}-1epoch+earlystop                 & xx.x & xx.x & xx.x & xx.x & \textasciitilde0.39s & 3B \\
% \DiscoSGQwenHalfB{}-1epoch+earlystop                & 50.8 & 91.0 & xx.x & xx.x & \textasciitilde0.39s & 0.5B \\
% \DiscoSGQwenOneFiveB{}-1epoch+earlystop             & 54.2 & 91.6 & xx.x & xx.x & \textasciitilde0.77s & 1.5B \\
% \DiscoSGQwenSevenB{}-1epoch+earlystop                 & xx.x & xx.x & xx.x & xx.x & \textasciitilde3.27s & 7B \\
% \midrule
&\DiscoSGTFiveBase{}                      & 52.7 & 91.7 & 38.4 & 88.5 & \textasciitilde0.07s & 0.25B \\
&\DiscoSGTFiveLarge{} (8730)                     & 69.4 & 95.1 & 53.0 & 91.8 & \textasciitilde0.32s & 0.78B \\
&\DiscoSGTFiveXL{}                 & \underline{76.8} & \underline{96.8} & 66.0 & 94.9 & \textasciitilde0.39s & 3B \\
&\DiscoSGQwenHalfB{}                & 54.2 & 90.1 & 48.5 & 90.0 & \textasciitilde0.39s & 0.5B \\
&\DiscoSGQwenOneFiveB{}             & 65.2 & 94.3 & 51.6 & 89.5 & \textasciitilde0.77s & 1.5B \\
&\DiscoSGQwenSevenB{}                 & 20.8 & 75.0 & 47.3 & 90.6 & \textasciitilde3.27s & 7B \\

\arrayrulecolor{black!30}\hline\arrayrulecolor{black}
% \multicolumn{5}{l}{\textbf{\DiscoSGDS{} (300 examples)}} \\
% \midrule
\multirow{2}{*}{\textbf{\begin{tabular}[c]{@{}l@{}}\large End-to-End (Discourse)\\ \textit{\textbf{\DiscoSGDS{} (300 examples)}}\end{tabular}}}
&\DiscoSGTFiveLarge{} (300)                      & 45.4 & 90.1 & 34.2 & 87.8 & \textasciitilde0.71s & 0.78B \\
% &\DiscoSGTFiveLarge{} (Overlap+Shifting Aug.)                      & 46.3 & 90.3 & 44.3 & 90.4 & \textasciitilde0.71s & 0.78B \\
% &\DiscoSGRefinerLarge{}                      & 52.1 & 91.8 &  &  & \textasciitilde0.71s & 1.03B \\
&\DiscoSGGPT{} (Teacher)                      & 73.1 & 96.0 & \underline{74.5} & \underline{96.4} & \textasciitilde36.9s & - \\
\midrule
% \multicolumn{7}{l}{\textbf{\large Few-Shot Prompting (3-shot)}} \\
\multirow{5}{*}{\textbf{\begin{tabular}[c]{@{}l@{}}\large Few-Shot Prompting\\ \textit{\textbf{(3-shot)}}\end{tabular}}}
&\QwenInstruct{}                         & 48.0 & 90.2 & 50.1 & 91.2 & \textasciitilde10.4s & 72B \\
&\LlamaInstruct{}                       & 52.0 & 91.6 & 53.5 & 92.4 & \textasciitilde14.1s & 70B \\
&\GPTFourText{}                   & 53.2 & 91.7 & 52.5 & 92.0 & \textasciitilde19.8s & - \\
&\GPTFourMulti{}                  & 55.6 & 92.3 & 54.4 & 92.4 & \textasciitilde33.2s & - \\
&\GraphRAG{}-\GPTFourText{}                                & 47.7 & 90.9 & 41.4 & 90.0 & \textasciitilde71.7s & - \\
\midrule
% \multicolumn{7}{l}{\textbf{\large Iterative Refinement}} \\
% \arrayrulecolor{black!30}\hline\arrayrulecolor{black}
\multirow{3}{*}{\textbf{\begin{tabular}[c]{@{}l@{}}\large Iterative Refinement\\ \textit{\textbf{Prior Methods}}\end{tabular}}}
% \multicolumn{7}{l}{\textbf{Prior Methods}} \\
&\PiVe                                 & 54.0 & 92.0 & 52.1 & 92.0 & \textasciitilde60.8s & - \\
&\SelfRefine                          & 37.2 & 86.2 & 34.6 & 83.5 & \textasciitilde99.0s & - \\
&\ProgRefine                          & 53.4 & 90.9 & 52.5 & 92.1 & \textasciitilde60.8s & - \\
\arrayrulecolor{black!30}\hline\arrayrulecolor{black}
% \rowcolor{gray!15}
% \midrule
% \multicolumn{7}{l}{\textbf{Our Method}} \\
% \rowcolor{gray!15}
% \multirow{3}{*}{\begin{tabular}[c]{@{}l@{}}\large \textbf{Iterative Refinement} \\ \textit{\textbf{Our Method}}\end{tabular}}
\multirow{3}{*}{\textbf{\begin{tabular}[c]{@{}l@{}}\large Iterative Refinement\\ \textit{\textbf{Our Method}}\end{tabular}}}
& \cellcolor{gray!15}\DiscoSGRefinerBase{}  & \cellcolor{gray!15}64.3 & \cellcolor{gray!15}94.3 & \cellcolor{gray!15}67.3 & \cellcolor{gray!15}95.1 & \cellcolor{gray!15}\textasciitilde0.43s & \cellcolor{gray!15}0.5B \\
& \cellcolor{gray!15}\DiscoSGRefinerLarge{} & \cellcolor{gray!15}66.2 & \cellcolor{gray!15}94.7 & \cellcolor{gray!15}67.3 & \cellcolor{gray!15}95.2 & \cellcolor{gray!15}\textasciitilde0.71s & \cellcolor{gray!15}1.03B \\
& \cellcolor{gray!15}\DiscoSGRefinerXL{}    & \cellcolor{gray!15}\textbf{66.7} & \cellcolor{gray!15}\textbf{94.9} & \cellcolor{gray!15}\textbf{68.6} & \cellcolor{gray!15}\textbf{95.4} & \cellcolor{gray!15}\textasciitilde2.17s & \cellcolor{gray!15}3.25B \\
\bottomrule
\end{tabular}%
}
\vspace{-3mm}
\caption{Comparison of methods on \DiscoSGDS{} Random and Length test sets. \textbf{Bold} values denote the best results among our method variants, while \underline{underlined} values show overall top performance.}
\label{tab:results_comparison}
\end{table*}
\vspace{-3mm}
}

\begin{table}[!]
\centering
\renewcommand{\arraystretch}{1.1}
\setlength{\tabcolsep}{3pt}
\resizebox{\columnwidth}{!}{%
\begin{tabular}{lcccc|c}
\hline
& \textbf{Cross.\,$\downarrow$} & \textbf{Long.\,$\downarrow$} & \textbf{Impl.\,$\downarrow$} & \textbf{Graph.\,$\downarrow$} & \textbf{Avg.\,$\downarrow$} \\
\hline
\FACTUALTBaseSentMerge{}  & 61.0  & 97.0   & 77.0   & 93.0  & 82.00 \\
\arrayrulecolor{black!30}\hline\arrayrulecolor{black}
\DiscoSGTFiveBase{}       & 67.0  & 100.0   & 96.0   & 100.0  & 90.75 \\
\GPTFourText{}            & 36.0  & 81.0   & 83.0   & 65.0  & 66.25 \\
\DiscoSGGPT{} (Teacher)   & 28.0  & 86.0   & 76.0   & 81.0  & 67.75 \\
\rowcolor{gray!15}
\DiscoSGRefinerLarge{}    & 43.0  & 85.0   & 74.0   & 79.0  & 70.25 \\
\arrayrulecolor{black!30}\hline\arrayrulecolor{black}
\textbf{Average}          & 47.0  & 89.8   & 81.2   & 83.6  & 75.4  \\
\hline
\end{tabular}
}
\vspace{-2mm}
\caption{
Discourse error rates (\%) identified by \GPTFour{}-based annotation across four error categories: Cross-Sentence Coreference Resolution (Cross.), Long-Range Relational Dependency (Long.), Implicit Information Inference (Impl.), and Graph Coherence (Graph.). See human evaluation results in Appendix~\ref{app:error}.
}
\label{tab:error_rate_gpt}
\vspace{-3mm}
\end{table}

\paragraph{Methods Compared.} We compare \DiscoSGRefiner{} against 21 methods across 5 categories.  

\textbf{\textsc{(i)} Sentence Parsing \& Merging} methods parse each sentence in a multi-sentence caption independently, then merge the resulting scene graphs. The parsers include \StanfordParser{}~\cite{schuster2015generating}, as used in the original SPICE implementation~\cite{anderson2016spice}, as well as \VGTBaseSentMerge{}~\cite{sharifzadeh2022improving} and \FACTUALTBaseSentMerge{}~\cite{li2023factual}, which use \FlanTBase{} models fine-tuned on sentence-level data from VG and FACTUAL.  

\textbf{\textsc{(ii)} End-to-End (Sentence)} methods, including \VGTBaseDirect{} and \FACTUALTDirect{}, apply these sentence-trained T5 models directly to full discourse inputs without intermediate merging.

\textbf{\textsc{(iii)} End-to-End (Discourse)} methods include \DiscoSGTFive{} (base/large/xl) and \DiscoSGQwen{} (0.5B/1.5B/7B parameters), which are \FlanT{} and \Qwen{}~\cite{bai2023qwen} variants fine-tuned on the full \DiscoSGDS{} training set (8,730 examples). We also report \DiscoSG{}-GPT-4o (Teacher) and \DiscoSGTFive{}-large (300), i.e., GPT-4o and \FlanTLarge{} fine-tuned on 300 expert-curated examples only.

% \textbf{\textsc{(iv)} Few-Shot Prompting} baselines use large instruction-tuned PLMs, including \QwenInstruct{} and \LlamaInstruct{}~\cite{grattafiori2024llama}, and also \GPTFour{} (text-only and multimodal). Generation uses 3-shot prompts where each shot is a caption-scene graph pair from \DiscoSGDS{}. We also report a 3-shot \GPTFour{}-based \GraphRAG{}~\cite{edge2024local} variant, where each shot includes the caption, extracted objects and relations, and the reference graph to condition generation.

\textbf{\textsc{(iv)} Few-Shot Prompting} baselines use large instruction-tuned PLMs, including \QwenInstruct{}, \LlamaInstruct{}~\cite{grattafiori2024llama}, and \GPTFour{} (text-only and multimodal). We use 3-shot prompts, each a caption-graph pair from \DiscoSGDS{}. We also report a 3-shot \GPTFour{}-based \GraphRAG{}~\cite{edge2024local} variant, where each shot includes the caption, extracted objects and relations, and the reference graph.

\textbf{\textsc{(v)} Iterative Refinement} methods, including \PiVe{}~\cite{han2024pive}, \SelfRefine{}~\cite{madaan2303self}, and \ProgRefine{}~\cite{li2024improving}, all employ 3-shot \GPTFour{} as both Generator and Interpreter, differing only in their Programmer implementations. 

All baseline details are in~\Cref{app:baselines}.

\subsubsection{Results and Analysis}
\noindent\textbf{\DiscoSGDS{} often enhances discourse parsing, with gains depending on model capacity, architecture, and training strategy.}
As shown in \Cref{tab:results_comparison}, few-shot \LlamaInstruct{} reaches 52.0/53.5 SPICE on Random/Length, surpassing \FACTUALTBaseSentMerge{} (49.4/52.7), the strongest baseline \emph{without using \DiscoSGDS{}}. Larger gains come from fine-tuning: the encoder-decoder \DiscoSGTFiveXL{} (3B) attains 76.8/66.0 SPICE (+55\% over 49.4 on Random; +25\% over 52.7 on Length), the decoder-only \DiscoSG{}-\GPTFour{} achieves 73.1/74.5 (+48\%/+41\% over the respective baselines), and our \DiscoSGRefinerBase{} (two fine-tuned \FlanTBase{} models) yields 64.3/67.3 SPICE, about \textbf{+30\%} over the sentence-merging baselines on both splits.

% \noindent\textbf{\DiscoSGDS{} often enhances discourse parsing, with gains depending on model capacity, architecture, and training strategy.}
% As shown in \Cref{tab:results_comparison}, few-shot \LlamaInstruct{} reaches 52.0/53.5 SPICE on Random/Length, surpassing \FACTUALTBaseSentMerge{} (49.4/52.7), the strongest non-\DiscoSGDS{} baseline. Larger gains come from fine-tuning: encoder–decoder \DiscoSGTFiveXL{} (3B) attains 76.8/66.0 (+55\% Random; +25\% Length), decoder-only \DiscoSG{}-\GPTFour{} achieves 73.1/74.5 (+48\%/+41\%), and our \DiscoSGRefinerBase{} (two fine-tuned \FlanTBase{} models) yields 64.3/67.3, about \textbf{+30\%} over the sentence-merging baselines on both splits.

However, discourse-level supervision alone is insufficient: robust performance across graph complexities requires either higher-capacity PLMs of the right architecture \emph{with} fine-tuning, or specialised procedures such as our iterative refinement. For example, within encoder-decoder models, scaling with fine-tuning yields the largest improvements, whereas naive end-to-end fine-tuning of small PLMs generalises poorly and is sensitive to graph complexity: \DiscoSGTFive{}-base (0.25B) and -large (0.78B) score 52.7 and 69.4 on Random but drop to 38.4 and 53.0 on Length, respectively. Decoder-only fine-tuned PLMs also underperform on Length unless scaled to \GPTFour{}-level sizes. \DiscoSGQwen{} at 0.5B, 1.5B, and 7B yields SPICE scores of 48.5, 51.6, and 47.3, each below the 52.7 baseline on Length split.

\noindent\textbf{\DiscoSGRefiner{} achieves the best balance of performance, generalisation, and cost.}
\DiscoSGRefiner{} models of all sizes outperform all other baselines on \textit{Length}, including those requiring vastly more computational resources ($\geq$70B vs. our \textbf{max 3.25B} parameters), except for \DiscoSGGPT{}. Nevertheless, \DiscoSGRefinerLarge{} scores around 7 SPICE and 1.2 BSSPICE points below \DiscoSGGPT{} on both splits while being \textbf{50$\times$} faster, and our base version achieves an \textbf{86$\times$} speedup. Beyond speed, our approach offers significant cost benefits: running \DiscoSGGPT{} on image captioning benchmarks \DetailCaps{}~\cite{dong2024benchmarking} and \CapArena{}~\cite{cheng2025caparena} would cost an estimated \textbf{\$25-150} in API fees, whereas all versions of our approach can execute locally on a single RTX 4090, eliminating API costs and network latency. \DiscoSGGPT{} cannot be released due to licensing restrictions, while our models are open-source. Compared to similar-cost methods like \DiscoSGTFive{} variants, our approach demonstrates superior generalisation: \DiscoSGTFive{} models perform well on Random but suffer degradation on Length (11-16 SPICE point drops). In contrast, our models maintain high performance (64 to 68 SPICE) across both test splits.

\noindent\textbf{Cross-sentence coreference resolution is the least difficult among the discourse parsing challenges.}
\Cref{tab:error_rate_gpt} highlights clear differences across error types in parser outputs. Parsers achieve the lowest error rates regarding linking entities across sentences, ranging from 28\% to 61\%. In contrast, the other three categories prove far more difficult, with rates clustering around 80\% to 90\%. The small \DiscoSGTFiveBase{} model shows the highest errors in every category and even exceeds the sentence-level baseline, confirming that small PLMs require specialised designs for discourse tasks. Overall, existing methods identify coreferent entities reasonably well, yet still struggle to maintain relational consistency, recover implicit information, and preserve graph coherence.

\subsection{Evaluation on Downstream VLM Tasks}
\label{subsec:downstream_eval}
{\setlength{\abovedisplayskip}{3pt}
\setlength{\belowdisplayskip}{5pt}
\begin{table}[t]
\centering
\setlength{\tabcolsep}{2.5pt} % Reduce column spacing
\resizebox{\columnwidth}{!}{%
\begin{tabular}{lccccccc}
\toprule
\rowcolor{white}
& \multicolumn{2}{c}{\textbf{\DetailCaps{}}} & \multicolumn{3}{c}{\textbf{\CapArena{}}} & \textbf{\DFOIL} & \textbf{Avg.} \\
\textbf{Metric} & $\tau\,\uparrow$ & $\rho\,\uparrow$ & $\tau\,\uparrow$ & $\rho\,\uparrow$ & Acc.\,$\uparrow$ & Acc.\,$\uparrow$ & \textbf{Rank\,$\downarrow$} \\
\midrule
Token Length & 8.8 & 12.2 & \underline{58.2} & 71.0 & \underline{68.3} & 2.00 & 12.0 \\
\midrule
\multicolumn{7}{l}{\textbf{\large N-gram}} \\
BLEU-4 & 25.6 & 36.4 & 31.9 & 42.4 & 47.4 & 53.8 & 14.3 \\
METEOR & 27.7 & 39.4 & 56.0 & \underline{77.1} & 61.2 & 61.3 & 7.5 \\
ROUGE-L & 23.6 & 33.4 & 1.1 & -6.4 & 48.0 & 50.2 & 17.7 \\
CIDEr & 24.7 & 35.1 & -27.9 & -20.9 & 38.4 & 52.0 & 19.5 \\
\midrule
\multicolumn{7}{l}{\textbf{\large Embedding-based}} \\
RefCLIPScore & 30.4 & 44.6 & -45.1 & -57.4 & 32.5 & 37.5 & 19.0 \\
Polos & 34.8 & 53.3 & 36.3 & 42.0 & 47.9 & 75.5 & 10.8 \\
\midrule
\multicolumn{7}{l}{\textbf{\large VLM-based}} \\
FLEUR & -3.5 & -2.0 & 29.7 & 39.3 & 45.8 & 18.0 & 19.3 \\
GPT-4o Eval & - & - & 94.3 & 84.6 & 62.7 & 91.5 & - \\  % <-- trailing ampersand removed
\midrule
\multicolumn{7}{l}{\textbf{\large Graph-based}} \\
\arrayrulecolor{black!30}\hline\arrayrulecolor{black}
SPICE w/ & & & & & & & \\
\ \ \textsc{FACTUAL (S+M)}$^\dagger$ & 40.7 & 55.1 & 36.3 & 46.8 & 52.8 & \textbf{86.5} & 8.2 \\
\ \ \DiscoSGTFive{}-base & 25.4 & 35.2 & 23.1 & 36.7 & 37.7 & 42.0 & 17.8 \\
\ \ \DiscoSGTFive{}-large & 31.8 & 43.5 & 3.3 & 9.9 & 42.4 & 53.0 & 15.8 \\
\rowcolor{gray!15}
\ \ \DiscoSGRefiner{}-base & 41.9 & 56.6 & 38.5 & 47.3 & 53.0 & \textbf{86.5} & 6.7 \\
\rowcolor{gray!15}
\ \ \DiscoSGRefiner{}-large & \textbf{\underline{42.2}} & \textbf{56.9} & \textbf{40.7} & \textbf{50.8} & \textbf{53.5} & \textbf{86.5} & \textbf{5.8} \\
\arrayrulecolor{black!30}\hline\arrayrulecolor{black}
BSSPICE w/ & & & & & & & \\
\ \ FACTUAL (S+M)$^\dagger$ & 38.8 & 56.6 & 49.5 & 62.2 & 54.5 & 88.0 & 6.2 \\
\ \ \DiscoSGTFive{}-base & 25.2 & 36.3 & 34.1 & 45.5 & 47.1 & 49.5 & 15.3 \\
\ \ \DiscoSGTFive{}-large & 30.6 & 43.9 & 7.7 & 13.0 & 47.5 & 57.5 & 14.0 \\
\rowcolor{gray!15}
\ \ \DiscoSGRefiner{}-base & 40.0 & 58.0 & 51.6 & 66.2 & 55.2 & 87.5 & 5.0 \\
\rowcolor{gray!15}
\ \ \DiscoSGRefiner{}-large & \textbf{40.3} & \textbf{\underline{58.5}} & \textbf{53.8} & \textbf{70.5} & \textbf{55.4} & \textbf{\underline{89.0}} & \textbf{\underline{3.7}} \\
\arrayrulecolor{black!30}\hline\arrayrulecolor{black}
CAPTURE w/ & & & & & & & \\
\ \ \textsc{FACTUAL (S+M)}$^\dagger$ & 39.6 & 54.5 & 53.8 & 61.3 & 57.6 & 69.5 & 7.2 \\
\ \ \DiscoSGTFive{}-base & 18.0 & 26.0 & -3.3 & 5.1 & 45.1 & 43.5 & 19.8 \\
\ \ \DiscoSGTFive{}-large & 22.7 & 32.5 & -18.7 & -17.8 & 46.1 & 55.0 & 18.7 \\
\rowcolor{gray!15}
\ \ \DiscoSGRefiner{}-base & 40.7 & 55.9 & 51.6 & 60.9 & \textbf{58.4} & 71.0 & 6.2 \\
\rowcolor{gray!15}
\ \ \DiscoSGRefiner{}-large & \textbf{40.8} & \textbf{56.0} & \textbf{58.2} & \textbf{74.1} & 58.2 & \textbf{74.0} & \textbf{4.0} \\
\bottomrule
\end{tabular}%
}
\vspace{-3mm}
\caption{Downstream task evaluation results. $^\dagger$\FACTUALTBaseSentMerge{}.}
\label{tab:downstream_results}
\vspace{-5mm}
\end{table}
}

We evaluate the impact of different automated metrics on image captioning assessment and hallucination detection across three benchmark datasets.

%both comparing LLM captions to gold references via automated metrics.

\paragraph{Benchmarks.}
For \textbf{Image Captioning}, we use \CapArena{}~\cite{cheng2025caparena}, which includes 6,000 human-written reference captions, each compared against outputs from 14 VLMs based on human preferences, and used to rank VLM performance. \DetailCaps{}~\cite{dong2024benchmarking} provides 4,870 images with captions rated 1-5 by \GPTFour{}. For \textbf{Hallucination Detection}, we introduce \DFOIL{}, a new benchmark designed to evaluate how well different metrics detect hallucinations in discourse-level outputs from VLMs. Inspired by \FOIL{}~\cite{shekhar2017foil}, \DFOIL{} contains 200 multi-sentence captions from \SharedGPTV{} with subtle hallucinated entities or relations. Each hallucinated caption is paired with a minimally corrected version annotated by humans, along with a reference caption generated by \GPTFourOne{}. See~\Cref{app:dfoil} for collection details.

\paragraph{Metrics.} 
We evaluate using \textbf{graph-based metrics} including SPICE, BSSPICE, and CAPTURE~\cite{dong2024benchmarking}, which measure similarity between the scene graphs of candidate and reference captions. Specifically, we compare these metrics when using scene graphs parsed by \DiscoSGRefiner{} versus those from the \FACTUALTBaseSentMerge{} baseline. \textbf{Other metrics include:} RefCLIPScore~\cite{hessel2021clipscore} and POLOS~\cite{wada2024polos} (embedding-based), FLEUR~\cite{lee2024fleur} and zero-shot \GPTFour{} (VLM-based), BLEU-4~\cite{papineni2002bleu}, METEOR~\cite{banerjee2005meteor}, ROUGE-L~\cite{lin2004rouge}, and CIDEr~\cite{vedantam2015cider} (n-gram), and caption length (longer captions preferred).

\paragraph{Evaluation Setup.} 
For image captioning, we compute Kendall's~$\tau$ and Spearman's~$\rho$ correlations between metric scores and gold-standard judgments (\GPTFour{} ratings for \DetailCaps{}, human preferences for \CapArena{}). For \CapArena{}, we also report agreement accuracy (Acc.), defined as the percentage of metric predictions that match human preferences. For hallucination detection on \DFOIL{}, we report classification accuracy (Acc.) in distinguishing hallucinated captions from their corrected versions. Finally, we compute the average performance rank of each metric (excluding \GPTFour{} Eval) by ordering methods by correlation or accuracy across all evaluation scenarios.

\subsubsection{Results and Analysis}
\noindent\textbf{Scene graphs from \DiscoSGRefiner{} enhance graph-based metrics across tasks.}
Table~\ref{tab:downstream_results} shows that replacing \FACTUALTBaseSentMerge{} with \DiscoSGRefinerLarge{} improves SPICE, raising Kendall’s $\tau$ from 40.7 to 42.2 on \DetailCaps{} and from 36.3 to 40.7 on \CapArena{}. BSSPICE also improves, with $\tau$ rising from 38.8 to 40.3 on \DetailCaps{}, from 49.5 to 53.8 on \CapArena{}, and classification accuracy increasing from 88.0\% to 89.0\% on \DFOIL{}. CAPTURE increases from 53.8 to 58.2 in $\tau$ on \CapArena{} and from 69.5\% to 74.0\% in accuracy on \DFOIL{}. Even the smaller \DiscoSGRefinerBase{} surpasses the baseline in nearly all settings, and, on average, replacing the sentence-level parser with \DiscoSGRefiner{} improves the rank of each graph-based metric by about one to two positions. In contrast, replacing \FACTUALTBaseSentMerge{} with \DiscoSGTFive{} parsers reduces performance for \textbf{all} graph-based metrics across all tasks. These results indicate that our refinement architecture produces higher-quality scene graphs, which in turn enable more accurate evaluation of image captioning and hallucination detection.

\noindent\textbf{Graph-based metrics with \DiscoSGRefinerLarge{} achieve top performance among all automated metrics.} BSSPICE attains the best average rank of 3.7, closely followed by CAPTURE at 4.0. These results outperform commonly used metrics such as METEOR, with an average rank of 7.5, and RefCLIPScore, which ranks 19.0. On \DFOIL{}, BSSPICE achieves 89.0 percent accuracy, second only to \GPTFour{} Eval at 91.5 percent, while METEOR and RefCLIPScore reach only 61.3 and 37.5 percent, respectively. This indicates that graph-based metrics are more effective at detecting subtle semantic issues, such as incorrect entities or relations, which are often missed by n-gram or embedding-based metrics.

\subsection{Ablation Study}
{
\setlength{\abovedisplayskip}{3pt}
\setlength{\belowdisplayskip}{5pt}
\begin{table}[t]
\centering
\setlength{\tabcolsep}{2.5pt}
\resizebox{1\columnwidth}{!}{%
\begin{tabular}{lcccc}
\toprule
 & \multicolumn{2}{c}{\textbf{\DiscoSGDS{} (Random)}} & \multicolumn{2}{c}{\textbf{\DetailCaps{}}} \\
\textbf{System / Condition} & \textbf{SPICE}\,$\uparrow$ & \textbf{BSSPICE}\,$\uparrow$ & $\tau\,\uparrow$ & $\rho\,\uparrow$ \\
\midrule
\rowcolor{gray!15}
\DiscoSGRefinerLarge{} (Default) & 66.2 & 94.7 & 42.2 & 56.9 \\
\midrule
\multicolumn{5}{l}{\textbf{Ablation: Training–data source / augmentation}} \\
%\quad Expert-only \DiscoSGED{}-base 15$\times$           & 51.1 & 91.5 & 41.8 & 56.5 \\
\quad Expert-only (300) \DiscoSGED{} 15$\times$           & 52.1 & 91.8 & 42.1 & 56.9 \\
\quad Expert+Synth. (8730) \DiscoSGED{} 1$\times$           & 62.9 & 94.0 & 41.9 & 56.5 \\
\quad Expert+Synth. (8730) \DiscoSGED{} 5$\times$           & 64.4 & 94.3 & 41.8 & 56.5 \\
\midrule
\multicolumn{5}{l}{\textbf{Ablation: Refinement depth}} \\
\quad 1 iteration                                      & 63.5 & 94.2 & 42.1 & 56.8 \\
\quad 3 iterations                                     & 66.7 & 94.8 & 42.3 & 57.1 \\
\midrule
\multicolumn{5}{l}{\textbf{Ablation: Programmer edit modality}} \\
\quad Deletion-only                                   & 48.9 & 90.4 & 41.9 & 56.7 \\
\quad Insertion-only                                  & 53.4 & 92.1 & 42.1 & 56.7 \\
\midrule
\multicolumn{5}{l}{\textbf{Ablation: Action-generation architecture}} \\
\quad Monolithic (no disentangle)         & 51.9 & 91.6 & 41.7 & 55.4 \\
\bottomrule
\end{tabular}}
\vspace{-3mm}
\caption{Ablation studies on \DiscoSGRefinerLarge{}. 
}
\label{tab:ablation_study}
\vspace{-2mm}
\end{table}
}
%\vspace{-3mm} Default: full \DiscoSGDS{}, 15$\times$ augmentation, two iterations, and both edit types.

\paragraph{Evaluation Setup.} Table~\ref{tab:ablation_study} probes three design axes of \DiscoSGRefinerLarge{}.  
The \emph{Default} row derives \DiscoSGED{} from the \underline{full} \DiscoSGDS{} (300 expert\,+\,8{,}430 synthetic instances) with 15 $\times$ augmentation, executes \underline{two} refinement iterations, and employs a disentangled \emph{Programmer} (encoder-deletion, decoder-insertion).  
\emph{Each variant changes exactly one of these factors while holding the others fixed.}  
\textbf{\textsc{(i)}}~\textbf{Training-data source}: \emph{Expert-only} \DiscoSGED{}~\emph{15$\times$} removes the synthetic portion of \DiscoSGDS{}.  
\textbf{\textsc{(ii)}}~\textbf{Augmentation scale}: using the full \DiscoSGDS{} data, we reduce the number of edit instances by setting the augmentation multiplier to 1$\times$ or 5$\times$.  
\textbf{\textsc{(iii)}}~\textbf{Refinement depth}: we run a single or three iterations.  
\textbf{\textsc{(iv)}}~\textbf{Programmer edit modality}: we activate only deletion or only insertion edits.  
\textbf{\textsc{(v)}}~\textbf{Action-generation architecture}: a \emph{Monolithic Encoder-Decoder} variant replaces the disentangled design, using one decoder for both edit types.  
Results are reported with SPICE and BSSPICE on the \DiscoSGDS{} test, and SPICE's Kendall’s~$\tau$ / Spearman’s~$\rho$ on \DetailCaps{}.
\subsubsection{Results and Analysis}
\textbf{Synthetic data, augmentation scale, refinement depth, and disentangled edit actions are the main drivers of performance.} \textbf{\textsc{i)}} Using \emph{Expert-only \DiscoSGED{} 15$\times$} lowers SPICE by 14 points, confirming that synthetic examples in \DiscoSGDS{} add crucial graph coverage beyond what expert data provides. \textbf{\textsc{ii)}} Expanding augmentation from 1$\times$ to 15$\times$ yields steady yet diminishing SPICE gains (62.9 to 66.2) and marginal BSSPICE improvements, indicating that moderate edit operation expansion is already beneficial. \textbf{\textsc{iii)}} Increasing refinement iterations from 1 to 3 boosts all metrics, with 3 passes achieving the overall best scores. We default to 2 mainly for efficiency. \textbf{\textsc{iv)}} Activating only deletion or insertion edits reduces SPICE by 12 to 17 points, showing the importance of their combination for good graph correction. \textbf{\textsc{v)}} Replacing the disentangled Programmer with a monolithic encoder-decoder degrades performance, as generating long, combined edit sequences proves too challenging for \FlanTLarge{} decoder.

\section{Related Work}

\paragraph{Scene Graph Parsing Methods.}
Scene graph parsing has been explored in images~\cite{zellers2018neural,tang2020unbiased,xu2017scene,zhang2019graphical,cong2022reltr,li2022sgtr,im2024egtr,wu2025psg4dllm}, video~\cite{rodin2024action}, and multimodal inputs~\cite{wu2025usg}. Our work focuses on text-based parsing, where two primary strategies exist. The most common approach uses direct generation: encoder-decoder models are fine-tuned on datasets like VG~\cite{wang2018scene,choi-etal-2022-scene} and FACTUAL~\cite{li2023factual}, or PLMs are prompted with few-shot examples from benchmarks like TSGBench~\cite{yang2025llm}. The second strategy converts text into an intermediate representation, such as dependency parses or AMR, which is then transformed into a scene graph~\cite{schuster2015generating,anderson2016spice,choi-etal-2022-scene}. Existing datasets like FACTUAL enable strong sentence-level parsing but focus on isolated sentences, causing trained parsers to overlook discourse context.

\paragraph{Downstream Applications.}
Scene graphs have been successfully applied in vision-language tasks. They improve image retrieval~\cite{johnson2015image,andrews2019scene} and guide structured image caption generation~\cite{zhong2020comprehensive,zeng2024meacap}. Scene graphs also serve as a foundational representation for evaluating caption factual accuracy via metrics like SPICE~\cite{anderson2016spice}, SoftSPICE~\cite{li2023factual}, and CAPTURE~\cite{dong2024benchmarking}. They are also used for identifying and mitigating hallucinations in VLM outputs~\cite{yu2024hallucidoctor}.

% \section{Conclusion}
% We introduce \DiscoSG{}, a new task for discourse-level text scene graph parsing, together with \DiscoSGDS{}, a dataset specifically curated to support effective discourse-level parsing. \DiscoSGDS{} includes 8,830 VLM-generated captions annotated with semantically coherent scene graphs that capture key discourse phenomena. To achieve high performance at low cost, avoiding the expense of large fine-tuned PLMs while surpassing the limitations of smaller models, we propose \DiscoSGRefiner{}, an efficient iterative refinement framework that uses small PLMs for initial graph generation, then refines graphs through triple insertion and deletion to produce high-quality scene graphs. Our smallest model achieves a 30\% improvement in SPICE over the best sentence-level baseline while significantly reducing inference cost compared to the \GPTFour{} parser. We further demonstrate the framework's effectiveness on downstream vision-language tasks and introduce \DFOIL{}, a new benchmark for evaluating hallucination detection in discourse-level VLM outputs.

\section{Conclusion}
We introduce \DiscoSG{}, a new task for discourse-level text scene graph parsing, along with \DiscoSGDS{}, a dataset of 8,830 VLM-generated captions annotated with semantically coherent scene graphs by human experts and fine-tuned \GPTFour{}, where the graphs capture key discourse phenomena. To achieve high performance at low cost, we propose \DiscoSGRefiner{}, an efficient iterative refinement framework that uses small PLMs for initial graph generation and then refines graphs through triple insertion and deletion. Our smallest model achieves a 30\% SPICE improvement over the best sentence-level baseline on two test splits while significantly reducing inference costs compared to the \GPTFour{} parser. We demonstrate the framework's effectiveness on downstream vision-language tasks and introduce \DFOIL{}, a benchmark for evaluating hallucination detection in discourse-level VLM outputs.

% \clearpage
% \newpage
\section*{Limitations}
Despite including images in our dataset, our current approach operates in a text-only setting for discourse-level scene graph parsing. All parsing methods except \GPTFourMulti{} rely solely on textual descriptions without incorporating visual cues from the original images. While this presents an opportunity for future work, our evaluation of SOTA multimodal parsers like \GPTFourMulti{}, as well as findings from Universal Scene Graph Generation~\cite{wu2025usg}, suggest that current models incorporating additional visual information do not yield significant improvements over the best text-only approaches. This indicates that effectively leveraging multimodal information to boost text scene graph parsing still remains challenging.
\section*{Acknowledgments}
This work is a collaboration across Wuhan University, Monash University, and RMIT University. We thank Wuhan University for access to GPU resources. OpenAI API costs were primarily paid from Zhuang Li’s personal account.

\paragraph{Author Contributions}
\begin{itemize}[leftmargin=*]
\item \textbf{Zhuang Li (RMIT University; senior author).} Led the core idea and methodology; authored the main draft and edited most subsequent drafts; oversaw project execution and advised the student on this project (informal supervision); contributed to and oversaw data annotation (e.g., graph validation and student annotator training); conducted a small set of experiments; covered most API fees.
\item \textbf{Shaoqing Lin (Wuhan University; student author).} Led experimental investigation and performed most data annotation; contributed additional ideas and methodological details; contributed to manuscript drafting and review.
\item \textbf{Chong Teng (Wuhan University; corresponding author).} Managed project administration and correspondence; student’s supervisor; coordinated access to computational resources and GPU.
\item \textbf{Donghong Ji (Wuhan University).} Provided computational resources and GPU access; co-supervises the student.
\item \textbf{Lizhen Qu (Monash University).} Provided discussions and high-level suggestions.
\item \textbf{Fei Li (Wuhan University).} Co-supervises the student.
\end{itemize}

\bibliography{custom}

\clearpage
\newpage
\appendix
\section*{Appendix}

\section{Parsing Evaluation Details}

\subsection{Complete Baseline Descriptions}
\label{app:baselines}
We give detailed descriptions of every baseline used in our comparisons.

\paragraph{Sentence Parsing \& Merging Methods.}
These methods first parse each sentence in a multi-sentence caption with a sentence-level parser, then merge the resulting graphs by concatenating triples and removing duplicates. The parsers include \StanfordParser{}~\cite{schuster2015generating}, which converts dependency parses into scene graphs and underlies the original SPICE implementation~\cite{anderson2016spice}, as well as \VGTBaseSentMerge{} and \FACTUALTBaseSentMerge{}. The latter two use \VGTBase{}~\cite{sharifzadeh2022improving} and \FACTUALTBase{}~\cite{li2023factual}, respectively, both of which are \FlanTBase{} models fine-tuned on the \VG{} and \FACTUAL{} datasets, respectively.

\paragraph{End-to-End (Sentence) Methods.}
\VGTBaseDirect{} and \FACTUALTDirect{} apply \VGTBase{} and \FACTUALTBase{} directly to full captions without using the sentence-level parsing and graph merging pipeline.

\paragraph{End-to-End (Discourse) Methods.}
These models are fine-tuned on caption-graph pairs from \DiscoSGDS{}. \DiscoSGTFive{} variants are \FlanT{} models in base (0.25B), large (0.78B), and xl (3B) sizes, while \DiscoSGQwen{} variants are \Qwen{} models with 0.5B, 1.5B, and 7B parameters. Unless noted, all are trained on 8{,}730 examples combining human and synthetic data. We also include \DiscoSGTFive{}-large (300), trained on 300 expert curated examples to control for training size, and \DiscoSGGPT{}, a \GPTFour{} teacher fine-tuned on the same 300 examples used to generate the 8{,}730 synthetic instances.

% \paragraph{End-to-End (Discourse) Methods.}
% These models are fine-tuned on caption-graph pairs from \DiscoSGDS{}. \DiscoSGTFive{} variants are fine-tuned \FlanT{} models in base (0.25B), large (0.78B), and xl (3B) sizes, while \DiscoSGQwen{} variants are fine-tuned \Qwen{} models with 0.5B, 1.5B, and 7B parameters. All are PLMs fine-tuned on 8,730 examples combining manual and synthetic data. We also include \DiscoSGTFive{}‑large (300), trained only on 300 expert curated examples to control for training size, and \DiscoSGGPT{}, a \GPTFour{} teacher fine-tuned on the same 300 examples our teacher model for generating 8730 synthetic instances.
%\DiscoSGGPT{} is our teacher model for generating synthetic data, which is \GPTFour{} fine-tuned on 300 expert-annotated examples.

% \paragraph{Few-Shot Prompting (3-shot).}
% We evaluate large PLMs including \QwenInstruct{}, \LlamaInstruct{}, and \GPTFour{} in both text-only and multimodal versions. During the evaluation, each caption in the test set receives three in-context examples retrieved from \DiscoSGDS{} based on cosine similarities between TF-IDF vectors of the training captions and the test caption. We also evaluate a \GraphRAG{} variant with \GPTFour{} that first extracts entities and relations from captions and then conditions graph generation on these cues, using 3-shot prompts in which each shot consists of a \emph{caption plus its extracted entities and relations, paired with the reference graph}.

\paragraph{Few-Shot Prompting (3 shot).}
We evaluate large instruction tuned PLMs, including \QwenInstruct{}, \LlamaInstruct{}~\cite{grattafiori2024llama}, and \GPTFour{} (text-only and multimodal). For each test caption, generation is conditioned on three in-context examples formatted as caption-scene graph pairs; the examples are retrieved from \DiscoSGDS{} using cosine similarity of TF-IDF vectors. We also evaluate a \GraphRAG{} variant with \GPTFour{} (text) that first extracts entities and relations from captions and then conditions graph generation on these cues, using 3-shot prompts where each shot provides the caption, its extracted entities and relations, and the reference graph.

\paragraph{Iterative Refinement Methods.}
All iterative baselines use 3-shot \GPTFour{} as Generator and Interpreter, differing only in the Programmer component. The three examples for the Generator are retrieved from \DiscoSGDS{} and those for the Interpreter are retrieved from \DiscoSGED{} using the same TF-IDF cosine similarity retrieval method described above.
\begin{itemize}
\item \textbf{\PiVe{}}~\cite{han2024pive}: The Programmer is a \FlanT{} model fine-tuned on \DiscoSGED{} that proposes insertion actions only.
\item \textbf{\SelfRefine{}}~\cite{madaan2303self}: The Programmer is \GPTFour{} that generates self-feedback containing insertion and deletion suggestions based on three feedback examples retrieved from \DiscoSGED{} using cosine similarities between TF-IDF vectors of captions.
\item \textbf{\ProgRefine{}}~\cite{li2024improving}: The Programmer is a \FlanT{} model fine-tuned on \DiscoSGED{} that requires the decoder to directly predict both insertion and deletion actions, unlike other methods that use the encoder for deletion and decoder for insertion.
\end{itemize}
All methods run \textbf{two refinement iterations} for a fair comparison with our framework.

\subsection{Metric Definitions}
\label{app:metrics}
\noindent\textbf{SPICE}~\cite{anderson2016spice} evaluates the parsing performance by computing an F1 score for each caption based on the overlap between predicted triples ($T_{\text{pred}}$) and gold standard triples ($T_{\text{gold}}$). To better capture semantic equivalence, it uses both exact string and synonym matching for entity nodes within triples.
The score is the harmonic mean of precision (P) and recall (R), defined as:
\begin{align}
    \text{SPICE} &= \frac{2 \cdot P \cdot R}{P + R}, \quad \text{where:} \nonumber \\
    P &= \frac{|T_{\text{pred}} \cap T_{\text{gold}}|}{|T_{\text{pred}}|} \nonumber \\
    R &= \frac{|T_{\text{pred}} \cap T_{\text{gold}}|}{|T_{\text{gold}}|} \nonumber
\end{align}
\noindent\textbf{Bi-SoftSPICE (BSSPICE)} is a symmetric version of SoftSPICE~\cite{li2023factual} that replaces exact matching with semantic similarity. SoftSPICE computes cosine similarities between Sentence-BERT embeddings~\cite{reimers2019sentence} of triple phrases from the predicted ($G_{\text{pred}}$) and gold ($G_{\text{gold}}$) graphs, then aggregates these triple similarities into a final graph-level score.

BSSPICE is the harmonic mean of the forward ($S_{\text{pg}}$) and backward ($S_{\text{gp}}$) SoftSPICE scores:
\begin{align}
    S_{\text{pg}} &= \text{SoftSPICE}(G_{\text{pred}}, G_{\text{gold}}) \nonumber \\
    S_{\text{gp}} &= \text{SoftSPICE}(G_{\text{gold}}, G_{\text{pred}}) \nonumber \\
    \text{BSSPICE} &= \frac{2 \cdot S_{\text{pg}} \cdot S_{\text{gp}}}{S_{\text{pg}} + S_{\text{gp}}} \nonumber
\end{align}

\noindent\textbf{Inference Time (s)} is reported as the average runtime per sample, computed over 100 test captions. For open-source PLMs with fewer than 70 billion parameters, inference is performed on a single NVIDIA RTX 4090 GPU. For API-based models, the runtime is measured using OpenAI’s endpoints for \GPTFour{}-based models and Alibaba’s public endpoints for \LlamaInstruct{} and \QwenInstruct{}.

Although \DiscoSGGPT{} and \GPTFourText{} are both based on \GPTFour{}, \DiscoSGGPT{} has approximately twice the inference time of \GPTFourText{}. This is consistent with OpenAI’s documentation, which notes that fine-tuned models often experience increased latency due to being hosted on separate infrastructure. As we do not have access to the backend deployment configurations of these proprietary APIs, our measurements serve as an approximate but practical basis for comparing inference efficiency.

\subsection{Discourse Error Analysis Methodology}
\label{app:error}

\paragraph{Error Type Taxonomy.}
To systematically analyse the outputs of discourse-level scene graph parsers, we developed a structured error analysis methodology using \GPTFour{} as an automated annotator and a human annotator. \GPTFour{} is guided by a detailed prompt (see Appendix \ref{fig:Template prompt for discourse-level scene graph error analysis}). The \GPTFour{} and human annotators evaluate candidate scene graphs against ground-truth graphs and their source captions. The analysis focuses on four principal categories of discourse-level errors:

% The \input command for your table would go here
\begin{table*}[!]
\centering
\renewcommand{\arraystretch}{1.1}
\setlength{\tabcolsep}{5pt}
\resizebox{2\columnwidth}{!}{%
\begin{tabular}{lccccc|ccccc}
\hline
& \multicolumn{5}{c|}{\textbf{\GPTFour{}-Annotated}} & \multicolumn{5}{c}{\textbf{Human-Annotated}} \\
 & \textbf{Cross.\,$\downarrow$} & \textbf{Long.\,$\downarrow$} & \textbf{Impl.\,$\downarrow$} & \textbf{Graph.\,$\downarrow$} & \textbf{Avg.\,$\downarrow$} 
 & \textbf{Cross.\,$\downarrow$} & \textbf{Long.\,$\downarrow$} & \textbf{Impl.\,$\downarrow$} & \textbf{Graph.\,$\downarrow$} & \textbf{Avg.\,$\downarrow$} \\
\hline
% Human.        & 68.0  & 92.0  & 86.0   & 94.0  &  &  &  & \\
\FACTUALTBaseSentMerge{}     & 61.0   & 97.0   & 77.0   & 93.0 & 82.00 & 100.0 & 100.0 & 70.0 & 100.0 & 92.5 \\
\arrayrulecolor{black!30}\hline\arrayrulecolor{black}
\DiscoSGTFiveBase{}       & 67.0  & 100.0  & 96.0   & 100.0 & 90.75 & 100.0 & 100.0 & 100.0 & 100.0 & 100.0 \\
\GPTFourText{}  & 36.0   & 81.0   & 83.0   & 65.0 & 66.25 & 60.0 & 80.0 & 80.0 & 90.0 & 77.5 \\
\DiscoSGGPT{} (Teacher)       & 28.0   & 86.0   & 76.0  & 81.0 & 67.75 & 50.0 & 80.0 & 80.0 & 100.0 & 77.5 \\
\DiscoSGRefinerLarge{}       & 43.0   & 85.0   & 74.0   & 79.0 & 70.25 & 30.0 & 100.0 & 100.0 & 80.0 & 77.5 \\
\arrayrulecolor{black!30}\hline\arrayrulecolor{black}
\textbf{Average}  & 47.0 & 89.8 & 81.2 & 83.6 & 75.4 & 68.0  & 92.0  & 86.0 & 94.0 & 85.0 \\
\hline
\end{tabular}
}
\caption{
Discourse parsing error rates (\%) for each parsing method, comparing automated error analysis by \GPTFour{} (\textbf{left}) against manual annotation (\textbf{right}). Lower values indicate better performance. Error categories are abbreviated as follows: \textbf{Cross.} (Cross-Sentence Coreference), \textbf{Long.} (Long-Range Dependency), \textbf{Impl.} (Implicit Inference), and \textbf{Graph.} (Graph Coherence). The automated analysis used the prompt shown in Appendix \ref{fig:Template prompt for discourse-level scene graph error analysis}.
}
\label{tab:error_rate}
\end{table*}

\begin{itemize}[leftmargin=*]
    \item \textbf{Cross-Sentence Coreference Resolution:} Failure to correctly resolve anaphoric references, where an entity is mentioned across multiple sentences. \\
    \textit{Example: Not linking "a woman" in one sentence to "she" in a subsequent sentence.}

    \item \textbf{Long-Range Relational Dependency:} Omission or misidentification of relationships between entities that span different sentences. \\
    \textit{Example: A connection between two objects mentioned in separate sentences is missing from the graph.}

    \item \textbf{Implicit Information Inference:} Failure to find relationships or attributes that are logically implied by the combined context of multiple sentences but not explicitly stated. \\
    \textit{Example: Not inferring the triple (cat, near, window) from the explicit statements (cat, on, mat) and (mat, under, window).}

    \item \textbf{Graph Coherence:} The generation of a structurally flawed graph that is fragmented, inconsistent, or incomplete at the discourse level. \\
    \textit{Example: The final graph consists of disconnected subgraphs or omits entities and relations crucial for a unified representation.}
\end{itemize}

For our quantitative analysis, we calculate the prevalence of each error type across the dataset. Each graph is evaluated for the presence or absence of each error type. If one or more instances of a specific error type occur within a graph, that graph is marked as exhibiting that error type and the count for that error type is incremented by one. We then compute the percentage of graphs in the dataset that exhibit each error type.

\paragraph{Analysis.}
Our results reveal a consistent performance ranking across both automated and human evaluation methods. The \GPTFour{}-based annotations separate the models into two clear performance tiers. The top tier comprises \GPTFourText{}, \DiscoSGGPT{} as the teacher model, and \DiscoSGRefinerLarge{}, all of which achieve comparable average error rates in the range of 66\% to 70\%. The second tier includes \FACTUALTBaseSentMerge{}, which has an average error rate of 82.0\%, and \DiscoSGTFiveBase{}, which reaches 90.75\%. This reflects a performance gap of more than 12 percentage points between the two groups. The findings suggest that leveraging discourse-level data from \DiscoSGDS{} enhances parsing quality at the discourse level, although smaller PLMs require specialised architectures to remain competitive. Human annotation further confirmed this performance disparity, with evaluators identifying errors in 92.5\% of graphs produced by \FACTUALTBaseSentMerge{} and 100\% of graphs generated by \DiscoSGTFiveBase{}.

To validate our automated evaluation, we measured inter-annotator agreement between \GPTFour{} and human experts on a subset of 50 samples. The analysis yielded a Jaccard score of 70.9\% and an F1 score of 74.4\%, indicating substantial agreement and demonstrating that \GPTFour{} serves as a reliable proxy for human judgment. This validation enables scalable and reproducible assessment of discourse-level phenomena, which would otherwise be infeasible to evaluate manually at scale.

\subsection{Implementation Details}

%\paragraph{Experimental Setup.}
We implement all models using the Hugging Face Transformers\footnote{\url{https://huggingface.co/}} library and conduct experiments on a single NVIDIA A100 80GB or RTX 4090 GPU. To ensure fair inference time comparisons, all speed measurements for open-source models with fewer than 70 billion parameters are conducted exclusively on the RTX 4090 GPU. Our Programmer models are trained for 3 epochs with a batch size of 2. For the composite loss function, the deletion loss ($\mathcal{L}_{\text{delete}}$) and insertion loss ($\mathcal{L}_{\text{insert}}$) are weighted equally, using coefficients $\lambda_{\text{del}} = \lambda_{\text{ins}} = 0.5$.
%Model performance on both tasks is evaluated after each training epoch.

% \paragraph{Task-Specific Prompting.}
% To enable a single model to handle both editing tasks, we adopt standard instruction-tuning practices and design task-specific prompt templates. These textual instructions clearly indicate the intended tasks to the Programmer models, which are based on \FlanT{}, and are structured as follows:

% \begin{tcolorbox}[
%   enhanced,
%   colback=black!5,
%   colframe=black!70,
%   fonttitle=\bfseries,
%   title=Deletion Prediction Prompt
% ]
% \begin{verbatim}
% Delete Task:
% Caption: {caption}
% Candidate Graph:
% {triple_1, triple_2, ..., triple_n}
% \end{verbatim}
% \end{tcolorbox}

% \begin{tcolorbox}[
%   enhanced,
%   colback=black!5,
%   colframe=black!70,
%   fonttitle=\bfseries,
%   title=Insertion Generation Prompt
% ]
% \begin{verbatim}
% Insert Task:
% Caption: {caption}
% Corrupted Graph: 
% {triple_1, triple_2, ..., triple_n}
% \end{verbatim}
% \end{tcolorbox}

% \noindent In these templates, \texttt{\{caption\}} denotes the source text, while \texttt{\{triple\_1, ...\}} refers to a comma-separated list of scene graph triples.

%\paragraph{Ethical Considerations.}
Human annotators involved in the manual data labelling process were compensated at a rate commensurate with the average local salary to ensure fair labour practices.

\subsection{Case Study}
\begin{figure*}[htbp]
\centering
\begin{tcolorbox}[colframe=black!50!gray, colback=white, coltitle=white, fonttitle=\bfseries\small, fontupper=\small]
\textbf{Image:} \\
\begin{center}
    \includegraphics[width=0.6 \linewidth]{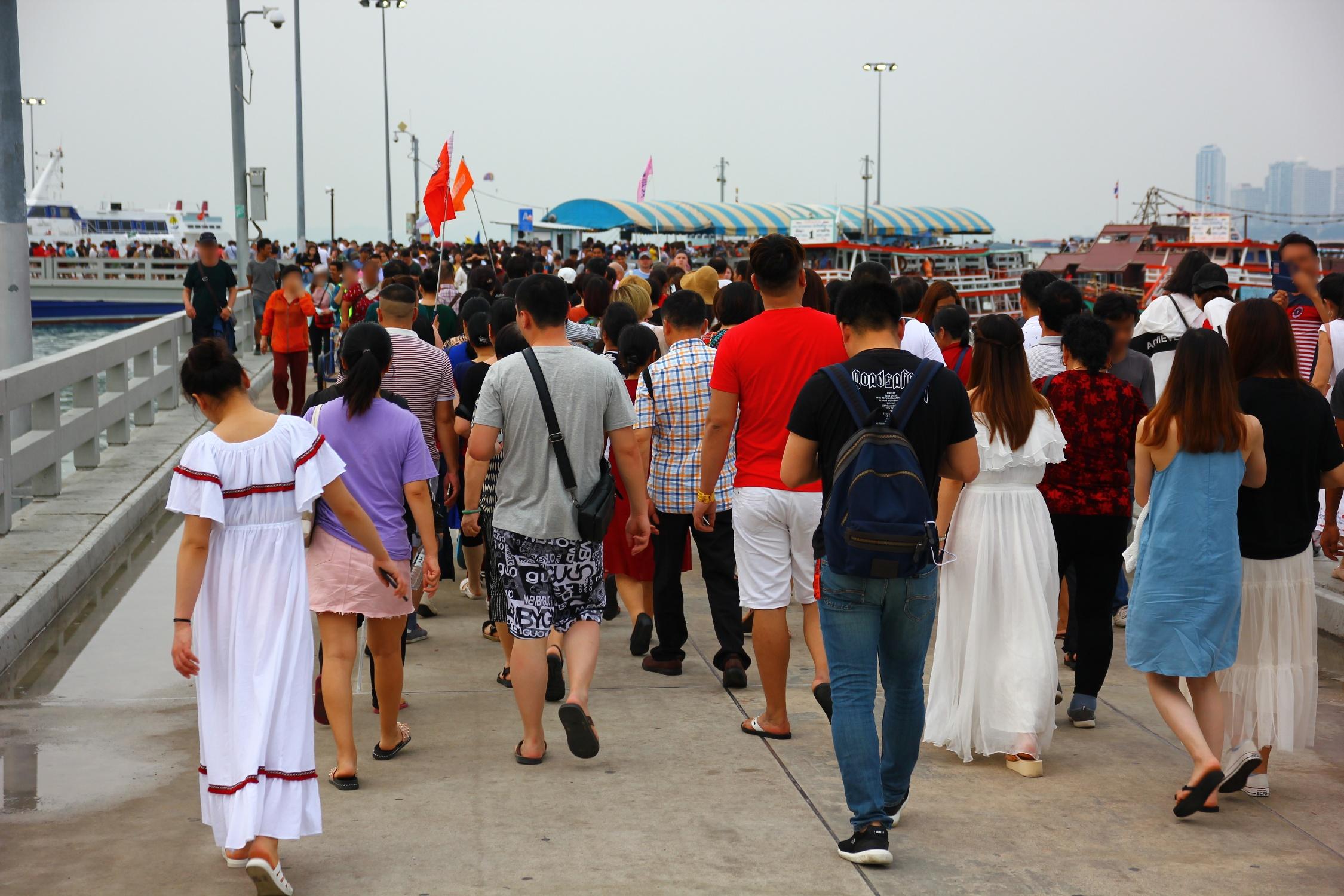} \\
\end{center}

\vspace{1.5em}

\textbf{Caption:} \\
In the image, a group of people are seen walking on a concrete pier towards a ferry terminal. The pier is equipped with a metal railing on the left side, providing safety for the pedestrians. The individuals are casually dressed, suitable for a summer day, and are carrying various bags and backpacks, suggesting they might be travelers or commuters.  Two red flags are flying on the left side of the image, possibly indicating some sort of warning or information. Further ahead, a blue and white canopy can be seen, likely providing shelter at the ferry terminal.  The sky above is hazy, creating a serene atmosphere. In the distance, tall buildings loom, indicating that the location is near a city or urban area. The image does not contain any discernible text.  The relative positions of the objects suggest a typical scene at a ferry terminal: people moving towards their destination, the safety measures in place, and the urban backdrop adding context to the setting. The image captures a moment of everyday life, with each element playing its part in the narrative. \\

\textbf{Initial Graph:} \\
( people , walk on , pier ) , ( people , walk towards , ferry terminal ) , ( pier , is , concrete ) , ( railing , at the left of , pier ) , ( railing , is , metal ) , ( individuals , carry , backpacks ) , ( individuals , carry , bags ) , ( individuals , is , casually dressed ) , ( flags , fly at the left of , image ) , ( flags , is , 2 ) , ( flags , is , red ) , ( canopy , at , ferry terminal ) , ( canopy , is , blue ) , ( canopy , is , white ) , ( sky , is , hazy ) , ( sky , is , serene ) , ( buildings , is , tall ) , ( buildings , near , city ) , ( image ) , ( backdrop , is , urban ) , ( objects , at , terminal ) , ( objects , in , setting ) , ( objects , is , relative ) , ( people , move towards , destination ) , ( terminal , is , ferry ) , ( element , play in , narrative ) , ( image , capture , narrative ) , ( narrative , is , everyday ) \\

\textbf{Deletion Prediction:} \\
( people , walk on , pier ) , ( people , walk towards , ferry terminal ) , ( pier , is , concrete ) , ( railing , at the left of , pier ) , ( railing , is , metal ) , ( individuals , carry , backpacks ) , ( individuals , carry , bags ) , ( individuals , is , casually dressed ) , ( flags , fly at the left of , image ) , ( flags , is , 2 ) , ( flags , is , red ) , ( canopy , at , ferry terminal ) , ( canopy , is , blue ) , ( canopy , is , white ) , ( sky , is , hazy ) , ( sky , is , serene ) , ( buildings , is , tall ) , ( buildings , near , city ) , {\color{red}\st{( image )}} , ( backdrop , is , urban ) , {\color{red}\st{( objects , at , terminal ) , ( objects , in , setting ) , ( objects , is , relative ) , ( people , move towards , destination ) , ( terminal , is , ferry ) , ( element , play in , narrative ) , ( image , capture , narrative ) , ( narrative , is , everyday )}} \\

\textbf{Insertion Generation:} \\
( railing , at the left of , pier ) , ( sky , is , hazy ) , ( people , walk on , pier ) , ( flags , is , 2 ) , ( individuals , carry , backpacks ) , {\color{darkgreen}( buildings , near , ferry terminal )} , ( pier , is , concrete ) , ( railing , is , metal ) , ( buildings , near , city ) , ( backdrop , is , urban ) , ( individuals , is , casually dressed ) , ( individuals , carry , bags ) , ( sky , is , serene ) , ( flags , fly at the left of , image ) , ( people , walk towards , ferry terminal ) , ( canopy , is , white ) , ( canopy , is , blue ) , ( canopy , at , ferry terminal ) , ( flags , is , red ) , ( buildings , is , tall ) , {\color{darkgreen}( people , is , group of )} \\

\end{tcolorbox}
\caption{Visualisation of the iterative scene graph refinement process in the \DiscoSGRefiner{} framework.} 
\label{fig:disco_case} 
\end{figure*}

We present a visual example to illustrate the iterative refinement process of the \DiscoSG framework. At the top, we show the input image (for reference only) alongside its multi-sentence textual description. The middle section displays the initial scene graph parsed from the description. At the bottom, we show the refined graph after applying the deletion and insertion modules, which remove redundant or irrelevant triples and add new, contextually appropriate ones. This example highlights how our framework enhances the accuracy and coherence of discourse-level scene graphs for complex image descriptions. See \Cref{fig:disco_case} for details.

\section{Data Collection Details}

\subsection{Active Learning Annotation for \DiscoSGDS{}}
\label{app:al_algo}
The pseudocode for our active learning annotation process is detailed in Algorithm~\ref{alg:active-learning}.
\begin{algorithm}[ht!]
\fontsize{10}{12}\selectfont
\caption{Active Learning Pipeline}
\label{alg:active-learning}
\KwIn{Seed training set $D_{\text{seed}}$ ($|D_{\text{seed}}| = 62$); fixed validation set $D_{\text{val}}$ ($|D_{\text{val}}| = 40$); unlabeled data pool $U$; number of iterations $N_{\text{iter}} = 2$; batch sizes per iteration $S = \{94,\,204\}$.}
\KwOut{Expanded training set $D_{\text{train}}$; sequence of fine tuned models $\{M_i\}_{i=0}^{N_{\text{iter}}}$; validation scores $\{P_i\}_{i=0}^{N_{\text{iter}}}$.}

$D_{\text{train}} \leftarrow D_{\text{seed}}$\;
$M_{0} \leftarrow \text{FineTune}(\text{GPT‑4o}, D_{\text{train}})$\;
$P_{0} \leftarrow \text{Evaluate}(M_{0}, D_{\text{val}})$\tcc*{initial SPICE score}

\For{$i \leftarrow 0$ \KwTo $N_{\text{iter}}-1$}{
    $B_{\text{raw}} \leftarrow \text{Sample}(S[i], U)$\;
    $U \leftarrow U \setminus B_{\text{raw}}$\;
    $B_{\text{draft}} \leftarrow \text{Predict}(M_{i}, B_{\text{raw}})$\;
    $B_{\text{refined}} \leftarrow \text{HumanRefine}(B_{\text{draft}})$\tcc*{two stage expert refinement}
    $D_{\text{train}} \leftarrow D_{\text{train}} \cup B_{\text{refined}}$\;
    $M_{i+1} \leftarrow \text{FineTune}(\text{GPT‑4o}, D_{\text{train}})$\;
    $P_{i+1} \leftarrow \text{Evaluate}(M_{i+1}, D_{\text{val}})$\;
}
\Return{$D_{\text{train}}, \{M_i\}_{i=0}^{N_{\text{iter}}}, \{P_i\}_{i=0}^{N_{\text{iter}}}$}
\end{algorithm}

\subsection{Annotation Guidelines for \DiscoSGDS{}}
\label{app:guidelines}

Annotation was conducted using Amazon Mechanical Turk's sandbox interface. Given the minimal risk associated with scene graph annotation tasks (extracting object relationships from visual descriptions), no additional risk disclaimers were required beyond standard research participation protocols.

To ensure the quality and consistency of our dataset, we established comprehensive guidelines for annotating scene graphs ($\vy$) from discourse descriptions ($\vx$) and images. The key principles and procedures are summarised below.

\paragraph{Core Task \& Principles.}
The primary objective is to extract a scene graph grounded in both the text and the image. This involves identifying object-attribute-value triples, such as \texttt{(mast, is, wooden)}, and object-relation-object triples, like \texttt{(mast, on, deck)}. All relations derived from verbs are explicitly marked with a \texttt{v:} prefix (e.g., \texttt{v:rest on}). The \texttt{v:} prefix is only applied during annotation to align with the FACTUAL-MR format~\cite{li2023factual} and is later removed. Only objects clearly visible or unambiguously inferable from the image are included as nodes. Abstract concepts (e.g., ``game''), viewpoint references (``camera''), and non-localizable elements are excluded.

\paragraph*{Annotation Workflow and Quality Control.}
The annotation process follows a structured workflow to maximise factual accuracy and consistency.
\begin{itemize}[leftmargin=*]
    \item \textbf{Hallucination Handling:} Annotators first cross-reference the text description ($\vx$) with the image to detect factual inconsistencies (hallucinations), such as incorrect object counts or non-existent relations. Instances with significant hallucinations that compromise core scene understanding are flagged and typically excluded from full annotation. Minor descriptive errors are corrected by the annotators.
    \item \textbf{Refinement Process:} Annotation is performed using a dedicated interface on Amazon Mechanical Turk\footnote{https://www.mturk.com/} where annotators refine initial draft graphs generated by \FACTUALTBaseSentMerge{} or \DiscoSGGPT{}. This involves adding, deleting, or modifying nodes (entities) and edges (relations).
    \item \textbf{Expert Review:} All manual annotations undergo a two-stage expert review that requires consensus for finalisation, ensuring high-quality outputs.
    \item \textbf{Intermediate Representation:} To reduce ambiguity during annotation, we use the FACTUAL-MR format~\cite{li2023factual}, which is then programmatically converted into the final scene graph structure.
\end{itemize}

\paragraph*{Semantic Representation Rules.}
Specific rules are applied to the representation of object attributes and relations:
\begin{itemize}[leftmargin=*]
    \item \textbf{Attributes:} Object attributes are represented using either the explicit form \texttt{(object, has\_attribute, attribute\_value)} or the shorthand form \texttt{(object, is, attribute\_value)}.
    \item \textbf{Relations:} Relations between objects or nodes follow the FACTUAL-MR convention, consisting of verbs and prepositions, with a preference for active voice where applicable. For example, \texttt{(court, v:separate by, lines)} is rewritten as \texttt{(lines, v:separate, court)} to ensure consistency, following the annotation guidelines of \FACTUAL{}~\cite{li2023factual}.
\end{itemize}

\paragraph*{Handling Discourse Phenomena.}
To address key discourse-level challenges such as Cross-Sentence Coreference, Long-Range Relational Dependencies, Implicit Inference, and Graph Coherence, our annotation guidelines emphasise the following procedures:

\begin{itemize}[leftmargin=*]
    \item \textbf{Entity Disambiguation and Coreference:} To resolve ambiguity, annotators create unique nodes for each distinct object, even if they share the same name. For instance, in ``Two cats are visible: one is sleeping, the other is playing,'' the cats are annotated as separate nodes (e.g., \texttt{cat:1}, \texttt{cat:2}). Similarly, collective nouns like ``dogs'' are disaggregated into individual nodes when described with different attributes or relations. For example, the phrase ``There are two dogs, one brown and one white'' results in distinct triples \texttt{(dog:1, is, brown)} and \texttt{(dog:2, is, white)}.

    \item \textbf{Quantifier Annotation:} Annotators capture the quantity of objects mentioned in the text (e.g., ``2 people'', ``several cats'') using a predefined set of labels (e.g., \texttt{1}, \texttt{2}, \texttt{many}, \texttt{uncountable}) associated with the corresponding entity nodes. To handle discourse-level counts, annotators aggregate these quantifiers across sentences. For instance, if one sentence mentions ``a cat'' and a later sentence mentions ``another cat,'' the final count for the ``cat'' entity is consolidated to the quantifier label \texttt{2}.

    \item \textbf{Implicit Relation Inference:} Annotators enrich the graph by inferring and adding relations that are logically implied by the context but not explicitly stated in a single sentence. For example, given the statements \texttt{(cat, on, mat)} and \texttt{(mat, under, window)}, the implicit spatial relation \texttt{(cat, near, window)} is added to the graph.

    \item \textbf{Entity Specificity and Semantic Precision:} Annotators are guided to use the most specific entity representation available (e.g., preferring ``cat'' over ``animal'') and to resolve semantic redundancy by merging nodes where appropriate (e.g., representing ``a husband and a wife'' as a single ``couple'' node).
\end{itemize}

%The full annotation guidelines are released alongside the dataset.

\subsection{Construction of the \DFOIL{} Dataset}
\label{app:dfoil}
The \DFOIL{} dataset was constructed through a multi-stage process designed to create a corpus of hallucinated captions, their corrected counterparts with minimal editing, and factually grounded reference captions.

\begin{enumerate}
    \item \textbf{Initial Data Collection:} We first gathered 410 image-text pairs where the captions were identified as containing discourse-level hallucinations. We also annotated the locations of these hallucinated entities or relations.
    
    \item \textbf{Reference Generation:} To establish a hallucination-free ground truth for each image, we employed \GPTFourOne{} to generate a new, detailed description. To maintain consistency with existing datasets, we used the same prompt as the \SharedGPTV{} dataset:
    
    \begin{tcolorbox}[
      enhanced,
      colback=black!5,
      colframe=black!70,
      fonttitle=\bfseries,
      title=Reference Generation Prompt
    ]
    Create detailed captions describing the contents of the given image. Include the object types and colours, counting the objects, object actions, precise object locations, texts, double-checking relative positions between objects, etc.
    \end{tcolorbox}
    
    These generated descriptions served as the factual references for the subsequent correction step.
    
    \item \textbf{Human Correction:} Using the generated references as a guide, our annotators manually revised and corrected a subset of 200 of the original hallucinated captions. This step ensured that the final corrected captions were factually aligned with the image content.
\end{enumerate}

Each finalised instance in the \DFOIL{} dataset comprises three key components: 1) the original caption containing hallucinations, 2) the corrected hallucination-free version of the hallucinated caption, and 3) the hallucination-free reference description generated by \GPTFourOne{}.

\subsection{Annotator Recruitment and Compensation}
\label{app:annotator_details}
Annotators were recruited from the research team based on their specialised expertise in scene graph parsing. All annotators were compensated through their regular institutional salaries with no additional per-annotation payments, ensuring fair labour practices. The recruitment of internal team members was necessary due to the specialised knowledge required for discourse-level scene graph annotation. To mitigate potential bias from this arrangement, we implemented rigorous quality control measures, including inter-annotator agreement assessment and two-stage expert review. %as described in Appendix~\ref{app:guidelines}.

\subsection{Dataset Licensing}
\label{app:licensing}

The \DiscoSGDS{} and \DFOIL{} datasets will be released under [MIT/Apache 2.0/CC BY 4.0] license for research purposes. The dataset includes expert annotations and synthesised examples that build upon existing research datasets, with appropriate consent obtained from research team members who participated in annotation. The dataset is intended for academic research use only and builds upon publicly available research datasets where consent was handled during original collection.

\section{Lexical Diversity Evaluation}
\label{sub:lexical}
\begin{table}[!]
\centering
\resizebox{\columnwidth}{!}{%
\begin{tabular}{lcc|cc}
\toprule
\multirow{2}{*}{Datasets} 
    & \multicolumn{2}{c}{Captions} 
    & \multicolumn{2}{c}{Graph} \\
 & MATTR $\uparrow$ & MTLD $\uparrow$
 & MATTR $\uparrow$ & MTLD $\uparrow$ \\
\midrule
% \DetailCaps{} & 0.61 & 0.73 & 57.12  & --   & --   & -- \\
% \CapArena{}  & 0.58 & 0.72 & 51.91  & --   & --   & -- \\
\VG{}        & 0.3682 & 13.5043 & 0.3563 & 12.5834 \\
\FACTUAL{}   & 0.3536 & 13.1511 & 0.3549 & 12.3893 \\
\TSGBench{} &  0.5168 & 9.8000 &  0.5137 & 9.2154 \\
\DiscoSGDS{} & 0.3809 & 14.1317 & 0.3596 & 13.1562  \\
%\DiscoSGED{}~\emph{1$\times$} & 0.59 & 0.75 & 58.93  & 0.51 & 0.58 & 20.38 \\
%\DiscoSGED{}~\emph{5$\times$} & 0.59 & 0.75 & 58.93  & 0.51 & 0.58 & 20.39 \\
\DiscoSGED{}~\emph{15$\times$} & 0.3809 & 14.1317 & 0.3603 & 12.4330  \\
\bottomrule
\end{tabular}%
}
\caption{Evaluation of lexical diversity in captions and scene graphs across different datasets.}
\label{tab:parser_faith_consistency}
\end{table}

To quantify linguistic diversity, we evaluate both the caption text and the linearised graph tokens of each dataset using two length-insensitive measures derived from the type-token ratio (TTR), the proportion of unique words (types) to total words (tokens). Because raw TTR declines with increasing text length, we adopt two widely accepted alternatives that offer more stable estimates:

\begin{itemize}[leftmargin=*]
    \item \textbf{Moving-Average TTR (MATTR)}~\cite{covington2010cutting}\,: computes the average TTR within a sliding window, reducing sensitivity to text length.
    \item \textbf{Measure of Textual Lexical Diversity (MTLD)}~\cite{mccarthy2005assessment}\,: measures the average number of tokens needed before the running TTR drops below a threshold, with higher scores indicating greater vocabulary diversity and lower redundancy.
\end{itemize}

To reduce the impact of varying average caption and graph token lengths across datasets, we compute diversity scores at the corpus level by concatenating all captions and all graph tokens, respectively, into single text sequences before computing diversity scores. This approach ensures a fair comparison across datasets. For both metrics, higher scores indicate greater lexical variety. Results are shown in Table~\ref{tab:parser_faith_consistency}.

\paragraph{Analysis.} Table~\ref{tab:parser_faith_consistency} shows that \DiscoSGDS{} achieves the highest MTLD scores, with 14.13 for captions and 13.16 for graph tokens, reflecting its strongest long-range lexical diversity among all datasets. While \TSGBench{} records the highest MATTR values, at 0.517 for captions and 0.514 for graphs, its much lower MTLD scores, 9.80 for captions and 9.22 for graphs, indicate that its lexical variety is not sustained over longer sequences. In contrast, \DiscoSGDS{} maintains a strong balance of both local and global diversity, making it particularly suitable for discourse-level modelling. The synthetic extension, \DiscoSGED{}, preserves this lexical richness, demonstrating that data augmentation does not diminish linguistic diversity.

\section{Prompt Descriptions}

\Cref{tab:prompt-templates} provides an overview of the prompt templates and example inputs used in our experiments. These templates standardize input formats and task-specific instructions across diverse scene graph parsing and caption analysis tasks, ensuring consistency in model prompting and evaluation.

\begin{table*}[!]
\setlength{\tabcolsep}{6pt}
\renewcommand{\arraystretch}{0.95}
\small
\centering
\begin{tabular}{p{0.22\textwidth} p{0.63\textwidth} p{0.1\textwidth}}
\toprule
\textbf{Template Name} & \textbf{Description} & \textbf{Link} \\
\midrule
\DiscoSGRefiner{} Deletion Prediction Template & Provides instructions for identifying and removing incorrect or irrelevant triplets from a candidate scene graph, based on a given caption. & \Cref{fig:Template prompt for DiscoSG Refiner Deletion Prediction} \\
\midrule
\DiscoSGRefiner{} Deletion Prediction Example & Shows a concrete input example used in the Deletion Prediction task. & \Cref{fig:Example prompt for DiscoSG Refiner Deletion Prediction} \\
\midrule
\DiscoSGRefiner{} Insertion Generation Template & Guides the model to add missing but contextually appropriate triplets to an incomplete scene graph, based on the caption and current graph. & \Cref{fig:Template prompt for DiscoSG Refiner Insertion Generation} \\
\midrule
\DiscoSGRefiner{} Insertion Generation Example & Provides an example input for the Insertion Generation task. & \Cref{fig:Example prompt for DiscoSG Refiner Insertion Generation} \\
\midrule
Hallucination Detection Template & Instructs the vision-language model to detect hallucinated content by comparing a caption to image content, producing a binary decision. This is used to filter data during synthetic data collection for \DiscoSGDS. & \Cref{fig:Template prompt for hallucination detection} \\
\midrule
\PiVe Prompt Template & Guides \GPTFour{} to generate initial scene graphs via few-shot prompting, then refine them using feedback from the \DiscoSGRefiner{} Programmer, conditioned on the caption. & \Cref{fig:Template prompt for Pive} \\
\midrule
\DFOIL{} Hallucination Detection Template & Prompts \GPTFour{} to determine which of two candidate captions is more semantically aligned with a reference caption, helping detect hallucinations. & \Cref{fig:Prompt Template for Discourse-Level Caption Hallucination Detection in D-FOIL} \\
\midrule
\SelfRefine{} Feedback Generation Template & Instructs \GPTFour{} to generate self-feedback on a scene graph, suggesting triplet insertions or deletions. & \Cref{fig:Template prompt for Self-Refine to generate Self-Feedback} \\
\midrule
\SelfRefine{} Refinement Template & Prompts \GPTFour{} to revise a previous scene graph based on generated self-feedback. & \Cref{fig:Template prompt for Self-Refine according to Self-Feedback} \\
\midrule
\ProgRefine{} Template & Applies a sequence of edits—insertions and deletions—predicted by \DiscoSGRefiner{} to update a scene graph. & \Cref{fig:Template prompt for Prog Refine} \\
\midrule
Discourse-Level Error Analysis Template & Prompts the model to analyze and classify scene graph errors by comparing with ground truth, focusing on coreference, relational dependencies, implicit inference, and coherence. & \Cref{fig:Template prompt for discourse-level scene graph error analysis} \\
\bottomrule
\end{tabular}
\caption{Descriptions of the prompt templates used in our experiments.}
\label{tab:prompt-templates}
\end{table*}

\begin{figure*}[htbp]
\centering
\begin{tcolorbox}[colframe=black!50!gray, colback=white, coltitle=white, fonttitle=\bfseries\small, fontupper=\small]

\textbf{Delete Task:} \\
\textbf{Caption:} \\
\{Caption\} \\
\textbf{Candidate Graph:} \\
\{Graph needs refine\}

\end{tcolorbox}
\caption{Prompt template for deletion prediction in \DiscoSGRefiner}
\label{fig:Template prompt for DiscoSG Refiner Deletion Prediction}
\end{figure*}

\begin{figure*}[htbp]
\centering
\begin{tcolorbox}[colframe=black!50!gray, colback=white, coltitle=white, fonttitle=\bfseries\small, fontupper=\small]
\textbf{Delete Task:} \\
\textbf{Caption:}
The image captures a serene scene in a park. A gravel path, dappled with sunlight filtering through the tall trees on either side, winds its way towards a white bridge. The bridge arches over a small body of water, possibly a stream or a pond. The sky above is a clear blue, with a few clouds scattered across it. The predominant colors in the image are the lush greens of the trees and grass, and the blue of the sky. The perspective of the image follows the path, leading the viewer's eye towards the bridge in the distance. The image exudes a sense of tranquility and invites one to take a leisurely stroll down the path. \\
\textbf{Candidate Graph:}
( path , is , gravel ) , ( trees , is , lush ) , ( trees , is , tall ) , ( path , wind towards , bridge ) , ( bridge , is , white ) , ( sky , is , gray ) , ( bridge , arch over , water ) , ( sky , is , clear ) , ( sky , is , blue ) , ( clouds , scatter across , sky ) , ( grass , is , lush ) , ( trees , on either side of , path ) , ( water , is , small ) , ( sunlight , dapple , path ) , ( sunlight , filter through , trees )

\end{tcolorbox}
\caption{Example input prompt for deletion prediction in \DiscoSGRefiner}
\label{fig:Example prompt for DiscoSG Refiner Deletion Prediction}
\end{figure*}

\begin{figure*}[htbp]
\centering
\begin{tcolorbox}[colframe=black!50!gray, colback=white, coltitle=white, fonttitle=\bfseries\small, fontupper=\small]

\textbf{Insert Task:} \\
\textbf{Caption:} \\
\{Caption\} \\
\textbf{Corrupted Graph:} \\
\{Graph needs refine / Graph refined by deletion prediction task\}

\end{tcolorbox}
\caption{Template input prompt for insertion generation in \DiscoSGRefiner}
\label{fig:Template prompt for DiscoSG Refiner Insertion Generation}
\end{figure*}

\begin{figure*}[htbp]
\centering
\begin{tcolorbox}[colframe=black!50!gray, colback=white, coltitle=white, fonttitle=\bfseries\small, fontupper=\small]
\textbf{Insert Task:} \\
\textbf{Caption:}
The image captures a serene scene in a park. A gravel path, dappled with sunlight filtering through the tall trees on either side, winds its way towards a white bridge. The bridge arches over a small body of water, possibly a stream or a pond. The sky above is a clear blue, with a few clouds scattered across it. The predominant colors in the image are the lush greens of the trees and grass, and the blue of the sky. The perspective of the image follows the path, leading the viewer's eye towards the bridge in the distance. The image exudes a sense of tranquility and invites one to take a leisurely stroll down the path. \\
\textbf{Corrupted Graph:}
( path , is , gravel ) , ( trees , is , lush ) , ( trees , is , tall ) , ( path , wind towards , bridge ) , ( bridge , is , white ) , {\color{red}\st{"( sky , is , gray )}Deleted by Deletion Prediction"} , ( bridge , arch over , water ) , ( sky , is , clear ) , ( sky , is , blue ) , ( clouds , scatter across , sky ) , ( grass , is , lush ) , ( trees , on either side of , path ) , ( water , is , small ) , ( sunlight , dapple , path ) , ( sunlight , filter through , trees )

\end{tcolorbox}
\caption{Example prompt for the insertion-generation step in \DiscoSGRefiner{}}
\label{fig:Example prompt for DiscoSG Refiner Insertion Generation}
\end{figure*}

\begin{figure*}[htbp]
\centering
\begin{tcolorbox}[colframe=black!50!gray, colback=white, coltitle=white, fonttitle=\bfseries\small, fontupper=\small]

\textbf{You are given an image and its description.} \\

\textbf{Analyse the description in detail to determine if it includes any hallucinations (information not present in the image).} \\

\textbf{Provide a detailed explanation of your reasoning.} \\

\textbf{At the end, provide a binary decision in the format:} \\

\textbf{"Final Decision: [Yes/No]".} \\

\textbf{Description:} \\
\{description\} \\
\{image upload here\}

\end{tcolorbox}
\caption{Prompt template for detecting hallucinations by comparing a caption to its corresponding image.}
\label{fig:Template prompt for hallucination detection} % 添加标签
\end{figure*}

\begin{figure*}[htbp]
\centering
\begin{tcolorbox}[colframe=black!50!gray, colback=white, coltitle=white, fonttitle=\bfseries\small, fontupper=\small]
\textbf{Transform the following text into a complete semantic graph and add the provided triple to the generated semantic graph.} \\

\textbf{Example 1:} \\

\textbf{Caption:} \{Caption\} \\
\textbf{Incomplete semantic graph:} \{incomplete scene graph\} \\
\textbf{Triple to add:} \{triple to add\} \\
\textbf{Complete semantic graph:} \{complete graph\} \\

\textbf{Example 2:} \\

\textbf{Caption:} \{Caption\} \\
\textbf{Incomplete semantic graph:} \{incomplete scene graph\} \\
\textbf{Triple to add:} \{triple to add\} \\
\textbf{Complete semantic graph:} \{complete graph\} \\

\textbf{Example 3:} \\

\textbf{Caption:} \{Caption\} \\
\textbf{Incomplete semantic graph:} \{incomplete scene graph\} \\
\textbf{Triple to add:} \{triple to add\} \\
\textbf{Complete semantic graph:} \{complete graph\} \\

\textbf{Now, transform the following caption into a complete semantic graph:} \\

\textbf{Caption:} {caption} \\

\textbf{Incomplete semantic graph:} \\
\{incomplete graph\} \\

\textbf{Triple to add:} \\
\{triple need to add\} {\color{blue} \# Generated by our Refiner} \\

\textbf{Complete semantic graph:}

\end{tcolorbox}
\caption{Prompt template used by \PiVe{} to complete a scene graph by adding a specified triplet.}
\label{fig:Template prompt for Pive} % 添加标签
\end{figure*}

\begin{figure*}[htbp]
\centering
\begin{tcolorbox}[colframe=black!50!gray, colback=white, coltitle=white, fonttitle=\bfseries\small, fontupper=\small]

\textbf{You are given a ground truth image caption and two candidate captions. Your task is to choose which candidate caption (Candidate 1 or Candidate 2) is closer in meaning and detail to the ground truth caption. Only output "Candidate 1" or "Candidate 2". Do not provide any explanation or analysis.} \\

\textbf{Ground truth caption:} \{ground truth\} \\
\textbf{Candidate 1:} \{cand1\} \\
\textbf{Candidate 2:} \{cand2\} \\

\textbf{Which candidate is closer to the ground truth caption?} \\

\end{tcolorbox}
\caption{Prompt template for identifying hallucinations in candidate captions by selecting the one that best aligns with the ground truth in \DFOIL.}
\label{fig:Prompt Template for Discourse-Level Caption Hallucination Detection in D-FOIL} % 添加标签
\end{figure*}

\begin{figure*}[htbp]
\centering
\begin{tcolorbox}[colframe=black!50!gray, colback=white, coltitle=white, fonttitle=\bfseries\small, fontupper=\small]
\textbf{You are an expert at analyzing scene graphs for accuracy and completeness. Your task is to evaluate a scene graph based on an image caption and suggest improvements.} \\

\textbf{A scene graph consists of triples in the format (subject, relation, object) that represent the entities and relationships in an image.} \\

\textbf{For the given caption and scene graph, identify:} \\

\textbf{1. Triples that need to be added (entities or relationships mentioned in the caption but missing from the graph)} \\

\textbf{2. Triples that need to be removed (incorrect, irrelevant, or redundant entries)
Here are some examples of how to provide refinement suggestions:}\\

\textbf{Example 1:} \\

\textbf{Caption:} \{caption\} \\
\textbf{Scene Graph:} \{scene graph\} \\
\textbf{insert triples:} \{triples need to insert\} \\
\textbf{delete triples:} \{triples need to delete\} \\

\textbf{Example 2:} \\

\textbf{Caption:} \{caption\} \\
\textbf{Scene Graph:} \{scene graph\} \\
\textbf{insert triples:} \{triples need to insert\} \\
\textbf{delete triples:} \{triples need to delete\} \\

\textbf{Example 3:} \\

\textbf{Caption:} \{caption\} \\
\textbf{Scene Graph:} \{scene graph\} \\
\textbf{insert triples:} \{triples need to insert\} \\
\textbf{delete triples:} \{triples need to delete\} \\

--- \\

\textbf{Now, please analyze this new case:} \\
\textbf{Caption:} \{caption\} \\
\textbf{Scene Graph:} \{scene graph\} \\

\textbf{Provide your recommendations in this format:} \\
\textbf{insert triples:} \{All triples that should be added\} \\
\textbf{delete triples:} \{All triples that should be removed\}
\end{tcolorbox}
\caption{Prompt template for \SelfRefine{} that guides \GPTFour{} to generate feedback by identifying insertions and deletions needed to improve a scene graph based on a caption.}

\label{fig:Template prompt for Self-Refine to generate Self-Feedback} % 添加标签
\end{figure*}

\begin{figure*}[htbp]
\centering
\begin{tcolorbox}[colframe=black!50!gray, colback=white, coltitle=white, fonttitle=\bfseries\small, fontupper=\small]
\textbf{Parse the following image description into a scene graph. A scene graph consists of triplets in the format (subject, relation, object).} \\

\textbf{I'll provide examples that show how to refine scene graphs based on feedback suggestions:} \\

\textbf{Example 1:} \\

\textbf{Caption:} \{caption\} \\
\textbf{Corrupted Graph:}
\{Corrupted scene graph\} \\
\textbf{Refinement Suggestion:} \\
\textbf{insert triples:} \{all triples that should be added\} \\
\textbf{delete triples:} \{all triples that should be removed\} \\
\textbf{Improved Scene Graph:} 
\{Target scene graph\} \\

\textbf{Example 2:} \\

\textbf{Caption:} \{caption\} \\
\textbf{Corrupted Graph:}
\{Corrupted scene graph\} \\
\textbf{Refinement Suggestion:} \\
\textbf{insert triples:} \{all triples that should be added\} \\
\textbf{delete triples:} \{all triples that should be removed\} \\
\textbf{Improved Scene Graph:} 
\{Target scene graph\} \\

\textbf{Example 3:} \\

\textbf{Caption:} \{caption\} \\
\textbf{Corrupted Graph:}
\{Corrupted scene graph\} \\
\textbf{Refinement Suggestion:} \\
\textbf{insert triples:} \{all triples that should be added\} \\
\textbf{delete triples:} \{all triples that should be removed\} \\
\textbf{Improved Scene Graph:} 
\{Target scene graph\} \\

\textbf{Now, generate an accurate scene graph for the following description:} \\

\textbf{Description:} \\
\{input\_caption\} \\

\textbf{Previous Graph:} \\
\{previous\_graph\} \\

\textbf{Refinement Suggestion:} \\
\{Self-Feedback from Self-Refine\} \\

\textbf{Generate an improved scene graph below:} \\

\end{tcolorbox}
\caption{Prompt template used in \SelfRefine for revising a scene graph according to \GPTFour{}-generated feedback.}
\label{fig:Template prompt for Self-Refine according to Self-Feedback} % 添加标签
\end{figure*}

\begin{figure*}[htbp]
\centering
\begin{tcolorbox}[colframe=black!50!gray, colback=white, coltitle=white, fonttitle=\bfseries\small, fontupper=\small]

\textbf{You are an expert at improving scene graphs based on edit actions. A scene graph consists of triples in the format (subject, relation, object).} \\

\textbf{Here are some examples of how to improve scene graphs based on edit actions:} \\

\textbf{Example 1:} \\

\textbf{Caption:} \{caption\} \\
\textbf{Scene Graph:} \{scene graph\} \\
\textbf{Edit Actions:} \\
\textbf{INSERT:} \{target triples need to insert\} \\
\textbf{DELETE:} \{target triples need to delete\} \\
\textbf{Improved Scene Graph:} \{original scene graph\} \\

\textbf{Example 2:} \\

\textbf{Caption:} \{caption\} \\
\textbf{Scene Graph:} \{scene graph\} \\
\textbf{Edit Actions:} \\
\textbf{INSERT:} \{target triples need to insert\} \\
\textbf{DELETE:} \{target triples need to delete\} \\
\textbf{Improved Scene Graph:} \{original scene graph\} \\

\textbf{Example 3:} \\

\textbf{Caption:} \{caption\} \\
\textbf{Scene Graph:} \{scene graph\} \\
\textbf{Edit Actions:} \\
\textbf{INSERT:} \{target triples need to insert\} \\
\textbf{DELETE:} \{target triples need to delete\} \\
\textbf{Improved Scene Graph:} \{original scene graph\} \\

---

\textbf{Now, please improve this new scene graph:} \\

\textbf{Caption:} \{caption\} \\

\textbf{Scene Graph:} \{scene graph\} \\

\textbf{Edit Actions:} \{programmer output\} \\

\textbf{Improved Scene Graph:}

\end{tcolorbox}
\caption{Prompt template used in \ProgRefine for applying edit actions to revise a scene graph.}
\label{fig:Template prompt for Prog Refine} % 添加标签
\end{figure*}

\begin{figure*}[htbp]
\centering
\begin{tcolorbox}[colframe=black!50!gray, colback=white, coltitle=white, fonttitle=\bfseries\small, fontupper=\small]

\textbf{Given the following inputs:} \\
\textbf{1. caption: A textual description of a scene} \\
\textbf{2. ground truth graph: The correct reference scene graph} \\
\textbf{3. candidate graph: The scene graph being evaluated} \\

\textbf{Analyze the scene graph parsing error by comparing the candidate graph against the ground truth graph and the caption. Determine which of the following error types best describes the primary issue. **For each error type, consider both errors of commission (incorrect outputs) and errors of omission (missing outputs that should be present).** Pay special attention to cases where the candidate graph fails to produce entities, links, or relations that are present in the ground truth graph or are required by the caption.} \\

\textbf{1. Cross-Sentence Coreference Resolution Error} \\
\textbf{Definition: Failure to correctly identify and link mentions of the same entity across different sentences, including both incorrect links and entirely missing coreference chains.} \\
\textbf{- Omission focus: Also count cases where coreference chains are completely absent because the parser failed to produce the necessary entities or links.} \\
\textbf{- Example: If "a woman" is mentioned in one sentence and later referred to as "she" or "the artist" in another, but the candidate graph either mislinks or fails to link these as the same entity, or does not produce the entities at all.}

\textbf{2. Long-Range Relational Dependency Error} \\
\textbf{Definition: Missing or incorrect relationships between entities mentioned in separate, potentially distant sentences.} \\
\textbf{- Omission focus: Count not only incorrect relations, but also missing long-range dependencies—i.e., when the candidate graph produces very few or no relations that should span across sentences.} \\
\textbf{- Example: If an object is introduced in one sentence and its relationship to another object is described in a different sentence, but the candidate graph fails to connect these entities, or omits the relationship entirely.}

\textbf{3. Implicit Information Inference Error} \\
\textbf{Definition: Failure to infer and represent relationships or attributes not explicitly stated but apparent from broader context.} \\
\textbf{- Omission focus: This category is inherently about omissions—specifically, failure to cover information that should be inferred based on the combined information in the caption and ground truth graph.} \\
\textbf{- Example: If the ground truth graph contains a relationship or attribute not textually explicit in the caption (e.g., inferring (cat, near, window) from "The cat is on the mat." and "The mat is under the window."), but the candidate graph omits this, it is an omission error.}

\textbf{4. Graph Coherence Error} \\
\textbf{Definition: Failures in producing a globally consistent representation of the entire scene, including contradictions, fragmentation, or incompleteness.} \\
\textbf{- Omission focus: Include cases of "graph incompleteness"—that is, when the overall graph lacks sufficient entities, links, or triplets to represent the complete scene described in the text.} \\
\textbf{- Example: The candidate graph only contains disconnected subgraphs or lacks key triplets, failing to provide a unified and comprehensive scene graph.}

\textbf{5. Others} \\
\textbf{Any error type that doesn't fit into the categories above.}

\textbf{---}

\textbf{For each error category that applies, do the following:} \\
\textbf{- State the category name as a section heading.} \\
\textbf{- Only consider the differences between the candidate graph and the ground truth graph, and between the candidate graph and the caption, strictly based on the provided data. Do not incorporate any information or assumptions beyond what is explicitly given in the data.} \\
\textbf{- Provide detailed reasoning, including:} \\
  \textbf{- Specific evidence from the graph (e.g., which triples or nodes are missing or illustrate the error).} \\
  \textbf{- An explanation of why this constitutes the given error type (refer to the definitions and examples above, including omission criteria).} \\
  \textbf{- An example or clarification if helpful.}

\textbf{---}

\textbf{At the end, under the heading "Applicable Error Categories", list all applicable error category labels (using the exact English names, separated by commas).} \\

\textbf{Inputs:} \\
\textbf{Caption:} \\
\{caption\}

\textbf{Ground Truth Graph:} \\
\{ground truth graph\}

\textbf{Candidate Graph:} \\
\{candidate graph\}

\end{tcolorbox}
\caption{Prompt template for analysing discourse-level scene graph errors, including coreference, long-range relations, implicit inference, and coherence, by comparing candidate graphs with ground truth graphs and captions.}
\label{fig:Template prompt for discourse-level scene graph error analysis} % 添加标签
\end{figure*}

% \section{Example Appendix}
% \label{sec:appendix}

% This is an appendix.

\end{document}